\documentclass[11pt]{article}

\usepackage{fullpage}

\usepackage{times}
\usepackage{soul}
\usepackage{url}
\usepackage[hidelinks]{hyperref}
\usepackage[utf8]{inputenc}
\usepackage[small]{caption}
\usepackage{graphicx}
\usepackage{amsmath}
\usepackage{amsthm}
\usepackage{booktabs}
\usepackage{algorithm}
\usepackage[noend]{algorithmic}
\usepackage{stmaryrd}
\usepackage{enumitem}
\usepackage{natbib}
\usepackage{doi}
\usepackage{tikz}
\usetikzlibrary{arrows,decorations.markings,arrows.meta}
\usepackage{hyperref}
\usepackage{amssymb}

\def\Sphere{{\cal S}}
\def\Srho{{\cal S}_\rho}
\def\Ssqrtd{{\cal S}\!\!_{\sqrt{d}}\,}

\def\z{{\bf z}}
\def\hz{{\bf \hspace*{0.03cm}z}}
\def\hs{$\hspace*{-0.02cm}$}

\def\N{M}

\newtheorem{theorem}{Theorem}

\newtheorem{lemma}{Lemma}

\title{\vspace*{-1.4cm} Eccentric Regularization: \\
Minimizing Hyperspherical Energy \\
without explicit projection
}

\author{{\bf Xuefeng Li and Alan Blair} \\
School of Computer Science and Engineering\\
University of New South Wales, Sydney, Australia\\
\texttt{\{xuefeng.li1,a.blair\}@unsw.edu.au} \\
}

\date{April 26, 2021; revised July 1, 2022}

\newcommand{\rpm}{\raisebox{.2ex}{$\scriptstyle\pm$}}

\begin{document}

\maketitle

% Authors must not appear in the submitted version. They should be hidden
% as long as the \iclrfinalcopy macro remains commented out below.
% Non-anonymous submissions will be rejected without review.

% The \author macro works with any number of authors. There are two commands
% used to separate the names and addresses of multiple authors: \And and \AND.
%
% Using \And between authors leaves it to \LaTeX{} to determine where to break
% the lines. Using \AND forces a linebreak at that point. So, if \LaTeX{}
% puts 3 of 4 authors names on the first line, and the last on the second
% line, try using \AND instead of \And before the third author name.

%\iclrfinalcopy % Uncomment for camera-ready version, but NOT for submission.

\begin{abstract}
Several regularization methods have recently been introduced which
force the latent activations of an autoencoder or
deep neural network to conform to either a Gaussian or hyperspherical
distribution, or to minimize the implicit rank of the distribution in
latent space.  In the present work, we introduce a
simple and novel regularizing
loss function which
simulates a
pairwise repulsive force between items and an
attractive force of each item toward the origin.
We show that minimizing this loss function in
isolation achieves a hyperspherical distribution,
and demonstrate its effectiveness as a regularizer for an image
auto\-encoder.
Moreover, a reduction in the regularization parameter
leads to a modest increase in the eccentricity
of the distribution in latent space.
This enhances image generation, and allows
the eigenvectors of the covariance matrix to be extracted as deep
principal components, which can be used for data analysis, image
generation, visualization and downstream classification.
\end{abstract}

\section{Introduction}
In recent years a number of regularization methods have been introduced which
force the latent activations of an auto\-encoder or deep neural
network to conform to either a hyper\-spherical or Gaussian distribution,
in order to encourage diversity in the latent vectors,
or to minimize the implicit rank of the distribution in latent
space.{\let\thefootnote\relax\footnote{{Accepted for publication in {\sc IJCNN} 2022. \\[0.1cm] This research was undertaken with the assistance of resources from the
National Computational Infrastructure (NCI Australia), an NCRIS enabled capability supported by the Australian Government.}}

Variational Autoencoders (VAE) \citep{DBLP:journals/corr/KingmaW13} and
related variational methods such as
\hbox{$\beta$-VAE} \citep{DBLP:conf/iclr/HigginsMPBGBML17}
force the latent
distribution to match a known prior distribution by minimizing the
Kullback-Leibler divergence. Normally, a
standard Gaussian distribution is used as the prior,
but alternatives such as the hyperspherical
distribution have also been investigated in the literature due to
certain advantages \citep{DBLP:conf/uai/DavidsonFCKT18}.
More recently, deterministic alternatives have been
proposed such as
Wasserstein Autoencoder (WAE) \citep{tolstikhin2018wasserstein},
VQ-VAE \citep{DBLP:conf/nips/OordVK17} and
RAE \citep{DBLP:conf/iclr/GhoshSVBS20}.

Several existing methods encourage diversity by maxim\-izing
pairwise dissimilarity between items,
drawing inspiration in part from a paper by
J.J.\ \cite{thomson1904xxiv} in which
various classical models are proposed for maintaining the electrons of
an atom in an appropriate formation around the nucleus.
Hyper\-spherical Energy Minimization \citep{liu2018learning}
has been used
to regularize the hidden unit activations in deep networks,
with a Thomson-like loss function which
projects each vector
onto a hypersphere and simulates a repulsive force
pushing these
projected items away from each other and spreading them evenly
around the sphere.
Other methods use an implicit projection by optimizing
for the cosine similarity between vectors
\citep{yu2011diversity,bao2013incoherent,xie2015diversifying}.

Recently, new compressive techniques have been developed with the aim of
minimizing the intrinsic dimension of the latent space.
This can be achieved by introducing
projection mappings \citep{lin2020regularizing}
or additional
linear layers between the encoder and decoder \citep{jing2020implicit},
or by other methods such as
optimizing for a variational lower bound on the mutual information between datapoints
\citep{grover2019uncertainty}.

The word \emph{eccentricity}
-- sometimes also called ``unevenness'' \citep{xie2017uncorrelation} --
can be used as a general term
for the extent to which the spectrum of the covariance matrix
in latent space
differs from that of a hyperspherical or standard normal distribution.
Methods such as VAE aim to give equal importance to all latent features,
while compressive techniques like \hbox{IRMAE} \citep{jing2020implicit}
explicitly encourage some
features to become dominant while others diminish.

In the present work, we introduce a simple and
novel regularizing loss function and show that it
is minimized when the latent vectors conform to
a hyperspherical distribution.
We demonstrate the effectiveness of this loss function
as a regularizer for an image autoencoder.
We find that, for our proposed method as well as for the
Wasserstein Autoencoder (WAE) \citep{tolstikhin2018wasserstein},
improved Fr\'echet Inception scores can be achieved
when the scaling parameter is reduced, allowing the eccentricity
of the distribution in latent space to moderately increase.
This increased eccentricity additionally enables the
eigenvectors of the covariance matrix to be
extracted as deep principal components,
which can be used for data analysis,
image generation, visualization and downstream classification.

%\begin{figure}[T]
%\includegraphics[width=0.48\textwidth]{percent.eps}
%\caption{Maximum$\,$percentage$\,$difference
%between$\,\rho\,$and$\,\sqrt{d}.\!\!\!\!$}
%\label{figure:percent}
%\end{figure}

%\begin{figure}[T]
%\includegraphics[width=0.5\textwidth]{force.eps}
%\includegraphics[width=0.52\textwidth]{force.png}
%\caption{Repulsive force between pairs of items,
%based on their distance, for dimension $d=64$.}
%\label{figure:force}
%\end{figure}

\section{Eccentric Loss Function}
\label{section:loss}

We consider a family of loss functions $\,l_{\mu,\N}\,$
on a set of $n$ items $\z_i\in{\mathbb R}^d$
of the form:
\[
l_{\mu,\N}(\{\z_i\}) = \frac{1}{n(n\!-\!1)}\!
\sum_{\;\z_i,\,\z_j}\!\!
K_{\mu,\N}(\z_i,\z_j),
\]

\vspace*{-0.4cm}\noindent
where
\[
K_{\mu,\N}(\z_i,\z_j) =
\bigl(\frac{||\hz_i||^2+||\hz_j||^2\!}{2}\bigr)
- \mu \N\log\bigr(1 + \frac{||\hz_i\!- \z_j||^2\!}{\!\!\N}\bigr).
\]
Note that the gradient of
$K_{\mu,\N}\,$
effectively simulates a repulsive force
between all pairs of items
$\z_i$,~$\z_j$,
equal to
$2\mu(\z_i\!-\!\z_j)/\bigl(1+||\z_i\!-\!\z_j||^2/\N\bigr)$,
and an attractive force
$(-\z_i)$
of each item $\z_i$ toward the origin.

\smallskip\noindent
Equation (1) ostensibly has two free parameters
$\mu$ and $\N$,
but we intend to set $\N$ as a function of
$\mu$ and the dimension~$d$
in such a way that the loss function
is minimized on a hyper\-spherical distribution of radius
approximately $\sqrt{d}$,
which is a close approximation
(in the sense of Wasserstein distance)
to the Standard Normal distribution.
The relationship between $\mu$, $\N$ and the radius $\rho$
of the stationary spherical distribution is determined
by this theorem:

\begin{theorem}
Let $S_\rho$ denote the uniform distribution on a hypersphere
of radius $\rho$ in dimension $d\ge 2$,
and let $\Gamma()$ denote the Gamma Function.
Then $S_\rho$ is a stationary point for $l_{\mu,\N}$,
provided the following expression is equal~to~$\frac{1}{\mu}\,$:
\[
\frac{\N}{\rho^2\!\sqrt{\pi}}
\frac{\Gamma(\frac{d}{2})}{\Gamma(\!\frac{d-1}{2}\!)}
\!\int_{u=1}^{1+\frac{4\rho^2}{\N}}
\!\frac{(\frac{\N}{2\rho^2}(u\!-\!1))^{\frac{d-1}{2}}
(2-\frac{\N}{2\rho^2}(u\!-\!1))^{\frac{d-3}{2}}\!\!\!}{u}\,du.
\]
\end{theorem}

\noindent
{\it Proof:}
(see Appendix)

\begin{minipage}{0.47\textwidth}
\begin{figure}[H]
\vspace*{-0.25cm}  
\hspace*{-0.5cm}
\includegraphics[width=1.03\textwidth]{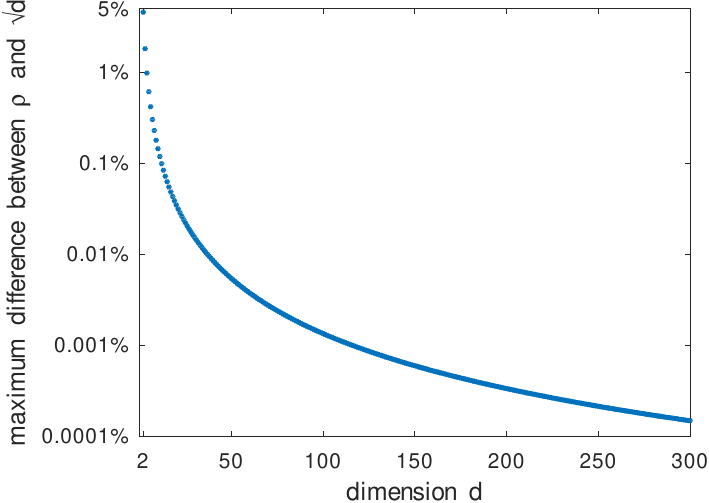}
\vspace*{0.1cm}  
\caption{Maximum percentage difference
between \\ $\rho$ and $\sqrt{d}$.}
%\caption{Maximum percentage difference
%between $\rho$ and $\sqrt{d}$.}
\label{figure:percent}
\end{figure}
%\vspace*{0.2cm}
\end{minipage}
\begin{minipage}{0.53\textwidth}
\begin{figure}[H]
\hspace*{-0.4cm}
\includegraphics[width=\textwidth]{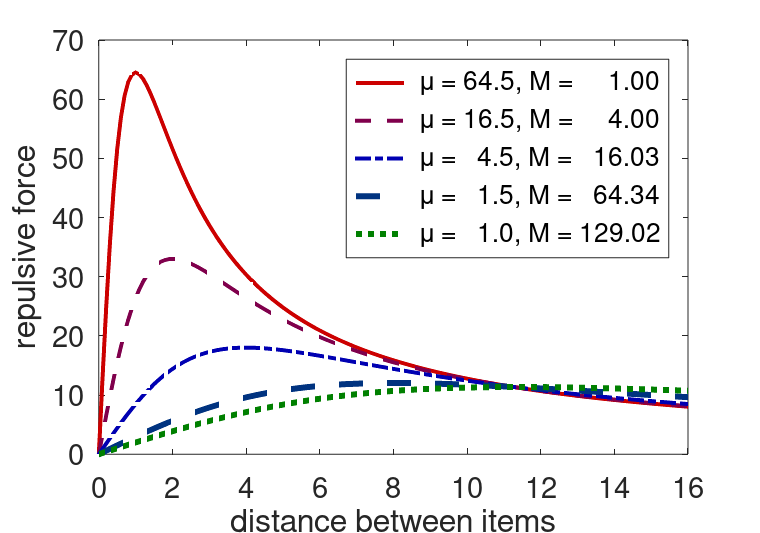}
\caption{Repulsive force between pairs of items, \\
based on their distance, for dimension $d=64$.}
\label{figure:force}
\end{figure}
\vspace*{0.2cm}
\end{minipage}

\bigskip\noindent
Our aim is to choose $\N$
in such a way that the radius $\rho$
of the stable spherical distribution
for $l_{\mu,\N}$
is very close to $\sqrt{d}\,$.
Henceforth we will use $l_\mu$ to denote $l_{\mu,\N}$ with $\N$
given by:
\[
\N=2d(1+\frac{1}{2\mu(d-1)})/(2\mu-1).
\]
(The motivation for this choice is explained in the Appendix).

\smallskip\noindent
In order to test the accuracy of this approximation,
we set $\N$ as above
and compute $\rho$ using numerical integration and bisection,
for each dimension $d$
and for all values of $\mu$ between $1$ and $2d+1$
(in increments of $0.01$).
Figure~\ref{figure:percent}
shows the percentage difference between
$\rho$ and $\sqrt{d}$, maximized over $\mu$,
for $d$ between $2$ and $300$.
We see that $\rho$ is within $0.1\%$ of $\sqrt{d}$ for $d\geq 12$,
within $0.01\%$ for $d\geq 38$, and within $0.001\%$ for $d\geq 117$.

%\begin{figure}[H]
%\includegraphics[width=0.48\textwidth]{percent.eps}
%\caption{Maximum$\,$percentage$\,$difference
%between$\,\rho\,$and$\,\sqrt{d}.\!\!\!\!$}
%\label{figure:percent}
%\end{figure}

%\begin{figure}[H]
%\includegraphics[width=0.52\textwidth]{force.png}
%\caption{Repulsive force between pairs of items,
%based on their distance, for dimension $d=64$.}
%\label{figure:force}
%\end{figure}

Figure~\ref{figure:force} shows
the magnitude of the
repulsive force between any two items based on the
distance between them, for various values of $\mu$
(and corresponding $\N$).
Note that the repulsive force increases until
a distance of $\sqrt{\N}$ and decreases thereafter.

\section{Regularization and Tolerance to Eccentricity}
\label{section:tolerance}

%We are primarily interested in the
We now focus on the use of this
eccentric loss function
for the purpose of regularization, as part of an overall loss function
of the form
\[
{\rm loss} = {\rm loss}_{\rm other} + \lambda\,l_{\mu}(\{\z_i\}).
\]
\smallskip\noindent
The scaling factor $\lambda$ and the parameter $\mu$
regulate both the local flexibility of the distribution
and its tolerance to eccentricity.

Figure~\ref{figure:force} shows
the magnitude of the
repulsive force between any two items based on the
distance between them.
%, for various values of $\mu$ and $\N$.
Note that the repulsive force increases until
a distance of $\sqrt{\N}$ and decreases thereafter.
At one extreme, where $\mu= d+\frac{1}{2}$ and $\N\simeq 1$,
the repulsive force is strongest for nearby items,
thus forcing them to spread out evenly on both a local and global scale.
At the other extreme, where $\mu=1$ and $\N\simeq 2d+1$,
each item is influenced most strongly by items that are far away,
%This ``action at a distance'' forces
thus forcing the items to spread out
on a global scale but allowing some flexibility on a local scale.

\begin{figure}[t]
\begin{center}
\begin{tabular}{ccc}
$\!\!\!\!\!\!$\includegraphics[width=0.32\textwidth]{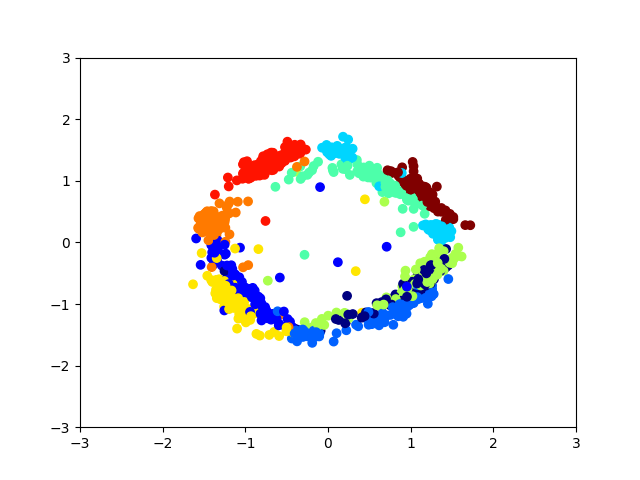}$\!\!\!\!\!\!$ &
$\!\!\!\!\!\!$\includegraphics[width=0.32\textwidth]{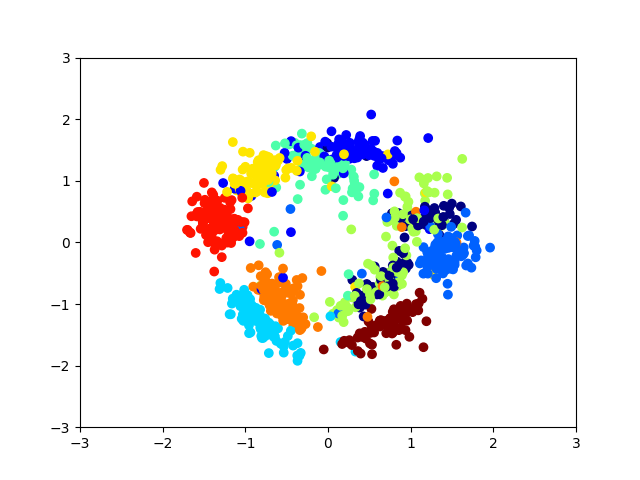}$\!\!\!\!\!\!$ &
$\!\!\!\!\!\!$\includegraphics[width=0.32\textwidth]{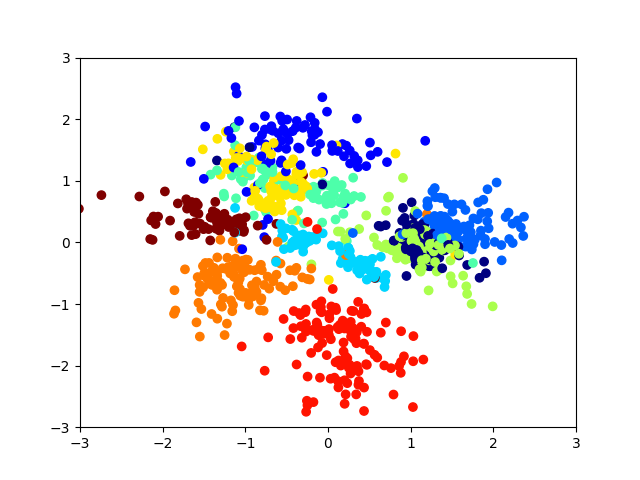}$\!\!\!\!\!\!$ \\[-0.1cm]
{\small (a) $\lambda = 0.1,\,\mu = 1.0$} &
{\small (b) $\lambda = 0.01,\,\mu = 1.0$} &
{\small (c) $\lambda = 0.001,\,\mu = 1.0$} \\
$\!\!\!\!\!\!$\includegraphics[width=0.32\textwidth]{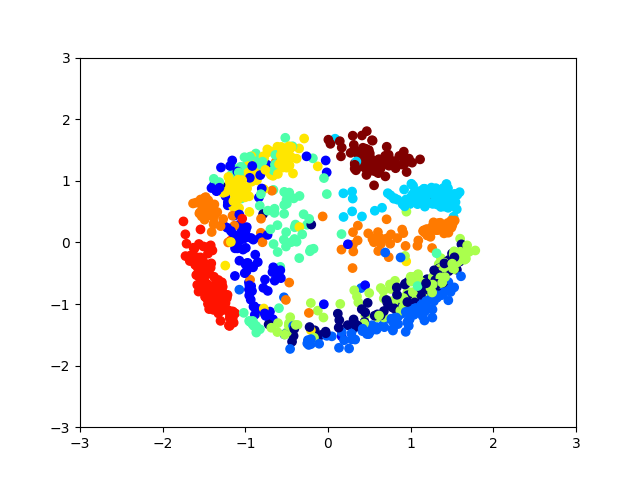}$\!\!\!\!\!\!$ &
$\!\!\!\!\!\!$\includegraphics[width=0.32\textwidth]{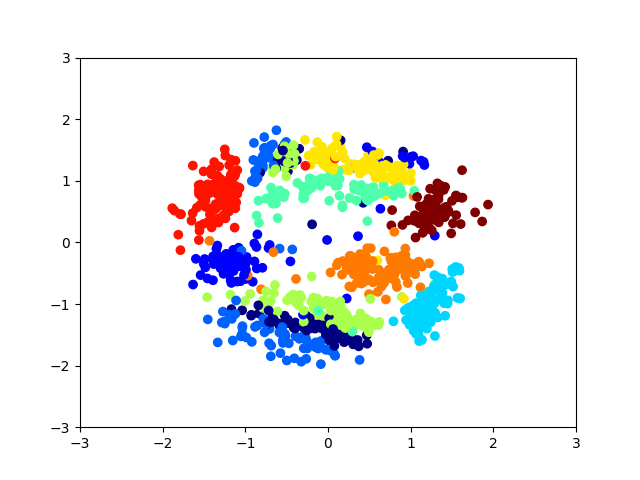}$\!\!\!\!\!\!$ &
$\!\!\!\!\!\!$\includegraphics[width=0.32\textwidth]{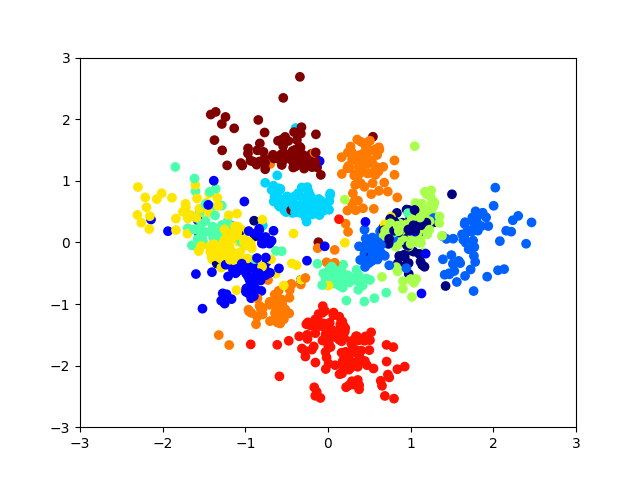}$\!\!\!\!\!\!$ \\[-0.1cm]
{\small (d) $\lambda=0.1,\,\mu=2.5$} &
{\small (e) $\lambda=0.01,\,\mu=2.5$} &
{\small (f) $\lambda=0.001,\,\mu=2.5$} \\[0.2cm]
\multicolumn{3}{c}{\includegraphics[width=0.45\textwidth]{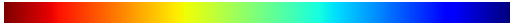}} \\
\multicolumn{3}{c}{0$\quad\;\;$1$\quad\;\;$2$\quad\;\;$3$\quad\;\;$4$\quad\;\;$5$\quad\;\;$6$\quad\;\;$7$\quad\;\;$8$\quad\;\;$9}
\end{tabular}
\caption{Distribution in latent space
of a random selection of test items,
for Eccentric Autoencoder trained
on the MNIST dataset using two latent dimensions,
with $\lambda=0.1$, $0.01$ and $0.001$;
$\mu=1.0$ and $2.5$.
%with different values of $\lambda$ and $\mu\,$.
The horizontal and vertical axes are the principal
components corresponding to the larger and smaller
eigenvalue of the covariance matrix.
}
\label{figure:mnist2_distribution}
\end{center}
\end{figure}

\subsection{Example: MNIST in Two Dimensions}

Figure~\ref{figure:mnist2_distribution} illustrates
the effect of eccentric regularization
in the case where
${\rm loss}_{\rm other}$
is the $L_2$ distance between the original and reconstructed images
for an autoencoder trained on the MNIST dataset, with $d=2$
(details are given in the next section).
When $\lambda$ is large (left column)
the distribution conforms closely
to a sphere of radius $\sqrt{d}$.
As $\lambda$ decreases (middle column)
the distribution deviates from the
sphere but starts to roughly approximate a
standard normal distribution.
When $\lambda$ decreases further (right column)
the eccentricity also increases,
allowing greater variance in some
directions than others.

\subsection{Intuitive Description}

We find experimentally that the eccentric loss function
constrains the trace of the covariance matrix to be
approximately equal to the dimension $d$.
This can be explained intuitively as follows:
Consider a point
$\z = (\sqrt{d},0,\ldots,0)$
on the sphere $\Ssqrtd$ of radius $\sqrt{d}\,$
(see Figure~\ref{fig:sphere_ellipsoid}).
Since the density of $\Ssqrtd$ is heavily concentrated in
vectors nearly perpendicular to $\z$,
we can think of a ``typical'' point $\z_\perp\in\Ssqrtd$
as being nearly perpendicular to $\z$ and exerting a repulsive force
on $\z$
whose component in the direction of $\z$ is approximately
\[
2\mu\,\z/\bigl(1+||\,\z-\!\z_\perp||^2/\N\bigr)
\simeq \z\bigl(2\mu/(1+2d/\N)\bigr)
\simeq \z\,.
\]
When these forces are aggregated,
the transverse components cancel each other out
while the radial components combine to exactly balance
the attractive force of $(-\z)$.
The influence of these ``typical'' points is particularly
dominant when $\mu\!=\!1$,
because in this case the pairwise repulsive force also reaches its
maximum at a distance of approximately
$\sqrt{2d}$ (see Figure~\ref{figure:force}).

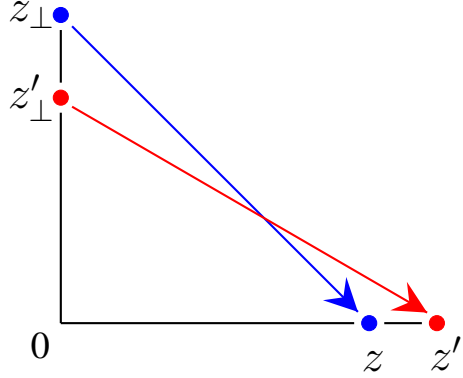
\begin{figure}[H]
\begin{center}
\vspace*{0.3cm}
\begin{tikzpicture}
  \draw[black, thick] (0,0) -- (0,2.8);
  \filldraw [red] (0,3) circle (0.1);
  \draw[black, thick] (0,3.2) -- (0,3.9);
  \filldraw [blue] (0,4.1) circle (0.1);

  \draw[black, thick] (0,0) -- (3.9,0);
  \filldraw [blue] (4.1,0) circle (0.1);
  \draw[black, thick] (4.3,0) -- (4.8,0);
  \filldraw [red] (5,0) circle (0.1);

  \draw [blue,thick,decoration={markings,mark=at position 1 with
    {\arrow[scale=3,>=stealth]{>}}},postaction={decorate}] (0.15,3.95) -- (3.95,0.15);

  \draw [red,thick,decoration={markings,mark=at position 1 with
    {\arrow[scale=3,>=stealth]{>}}},postaction={decorate}] (0.15,2.88) -- (4.87,0.14);

  \node[text width=0.4cm,black] at (-0.2,-0.3) {\Large 0};

  \node[text width=1cm] at (4.5,-0.5) {\LARGE $z$};
  \node[text width=1cm] at (5.4,-0.4) {\LARGE $z^\prime$};

  \node[text width=1cm] at (-0.2,4.1) {\LARGE $z_\perp$};
  \node[text width=1cm] at (-0.2,3.0) {\LARGE $z_\perp^\prime$};

\end{tikzpicture}
\vspace*{-0.3cm}
\end{center}
\caption{The point
$\z=(\sqrt{d},0,\ldots,0)$ is repelled by a
\hbox{``typical''}~point $\z_\perp$ on the sphere $\Ssqrtd$ with a force
whose \hbox{radial}~component approximately balances the
attractive force $(-\z)$;
when the sphere $\Ssqrtd$ is transformed to an
ellipse~with covariance
$\Sigma={\rm diag}(\sigma_1^2,\ldots,\sigma_d^2)$
with ${\rm Trace}(\Sigma)=\sum_i \sigma_i^2=d$,
the transformed points $\z'$, $\z'_\perp$ satisfy
$||\,\z'-\z'_\perp||\simeq||\,\z-\z_\perp||$
and the radial component continues to
approximate the (new) attractive force $(-\z')$.}
\label{fig:sphere_ellipsoid}
\end{figure}

Now suppose that the sphere $\Ssqrtd$ is
transformed into an
ellipse with covariance $\Sigma$,
%${\cal E}_\Sigma$, with
where
${\rm Trace}(\Sigma)=d$, and assume for simplicity that
$\Sigma={\rm diag}(\sigma_1^2,\ldots,\sigma_d^2)$~is~diagonal.
The transformation $\Sigma^{\frac{1}{2}}$
maps $\z$ to $\z'=(\sigma_1\sqrt{d},0,\ldots,0)$,
and the condition
$\sum_i\sigma_i^2={\rm Trace}(\Sigma)=d$
ensures that
the ``typical'' point $\z_\perp$ will be mapped to a point
$\z'_\perp$ for which
$||\,\z'-\z'_\perp||\simeq||\,\z-\z_\perp||$.
But, due to the change in angle,
the radial component of the repulsive force
will change from $(-\z)$ to $(-\z')$,
keeping it in balance with the attractive force.

\subsection{Minimum Loss Value}
When $\mu\!=\!1$,
we find experimentally that the minimum value
of the loss function $l_{\mu}^{\,d}$
obtainable through gradient descent
can be approximated by
\[
l_{\mu}^{\,d}\simeq d(1-2\log 2)\simeq -0.386\,d\,.
\]
This can be understood using the same intuition
as above, namely
that two random points
$\z_i, \z_j$ on $\Ssqrtd$
are likely to be nearly
perpendicular,
so an approximate value for
$l_{\mu}$ on $\Ssqrtd$ is given by:
\begin{align}
K_{\mu,\N}(\z_i,\z_j)
&=\bigl(\frac{||\hz_i||^2+||\hz_j||^2\!}{2}\bigr)
- \mu \N\log\bigr(1 + \frac{||\hz_i\!- \z_j||^2\!}{\!\!\N}\bigr) \nonumber \\
&\simeq\;\;\frac{d+d}{2}\;-\;(1)(2d\!+\!1)\log(1+\frac{2d}{2d+1}) \nonumber \\
&\simeq\;\; d(1-2\log 2). \nonumber
\end{align}

\hspace*{-0.5cm}
\begin{minipage}{0.48\textwidth}
\begin{figure}[H]
\vspace*{-0.6cm}
\hspace*{-0.4cm}
\includegraphics[width=1.14\textwidth]{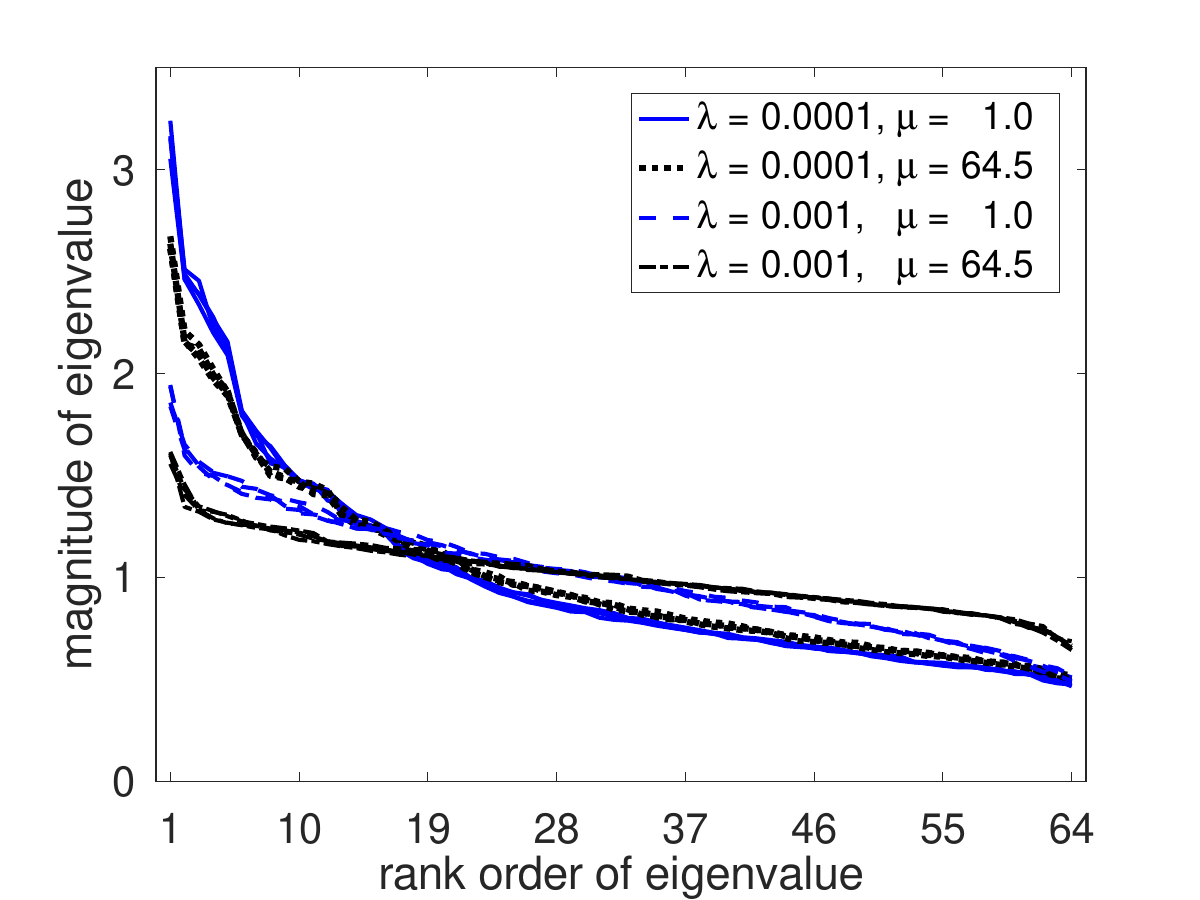}
%\vspace*{0.1cm}  
\caption{Spectrum of latent variable
\hbox{covariance} matrix for
Eccentric Autoencoders trained on CelebA;
four indepedent runs are superimposed for each
combination of $\lambda$ and $\mu$.}
\label{figure:spectrum}
\end{figure}
\end{minipage}
\begin{minipage}{0.04\textwidth}
\phantom{.}
\end{minipage}
\begin{minipage}{0.48\textwidth}

\begin{figure}[H]
\hspace*{-0.15cm}\begin{tabular}{|cccccccccc|}
\hline
 & & & & & &\multicolumn{1}{|c}{} & & & \\[-0.2cm]
  \!\!\includegraphics[width=0.09\textwidth]{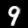}\!\!\! &
\!\!\!\includegraphics[width=0.09\textwidth]{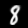}\!\!\! &
\!\!\!\includegraphics[width=0.09\textwidth]{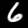}\!\!\! &
\!\!\!\includegraphics[width=0.09\textwidth]{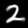}\!\!\! &
\!\!\!\includegraphics[width=0.09\textwidth]{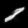}\!\!\! &
\!\!\!\includegraphics[width=0.09\textwidth]{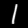}\!\! &
\multicolumn{1}{|c}{\!\!\includegraphics[width=0.09\textwidth]{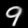}\!\!\!} &
\!\!\!\includegraphics[width=0.09\textwidth]{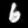}\!\!\! &
\!\!\!\includegraphics[width=0.09\textwidth]{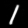}\!\!\! &
\!\!\!\includegraphics[width=0.09\textwidth]{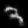}\!\! \\
  \!\!\includegraphics[width=0.09\textwidth]{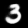}\!\!\! &
\!\!\!\includegraphics[width=0.09\textwidth]{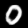}\!\!\! &
\!\!\!\includegraphics[width=0.09\textwidth]{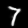}\!\!\! &
\!\!\!\includegraphics[width=0.09\textwidth]{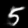}\!\!\! &
\!\!\!\includegraphics[width=0.09\textwidth]{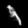}\!\!\! &
\!\!\!\includegraphics[width=0.09\textwidth]{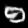}\!\! &
\multicolumn{1}{|c}{\!\!\includegraphics[width=0.09\textwidth]{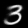}\!\!\!} &
\!\!\!\includegraphics[width=0.09\textwidth]{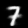}\!\!\! &
\!\!\!\includegraphics[width=0.09\textwidth]{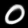}\!\!\! &
\!\!\!\includegraphics[width=0.09\textwidth]{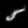}\!\! \\
 $p1$ & $p2$ & $p3$ & $p4$ & $p5$ & $p6$ &\multicolumn{1}{|c}{$p1$} & $p2$ & $p3$ & $p4$ \\[0.1cm]
\multicolumn{6}{|l}{$\qquad\qquad\qquad$(a) $\,d=6$} & \multicolumn{4}{|c|}{(b) $\,d=4$} \\[0.1cm]
\hline
 & & & & & & & &\multicolumn{1}{|c}{} & \\[-0.2cm]
  \!\!\includegraphics[width=0.09\textwidth]{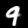}\!\!\! &
\!\!\!\includegraphics[width=0.09\textwidth]{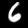}\!\!\! &
\!\!\!\includegraphics[width=0.09\textwidth]{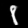}\!\!\! &
\!\!\!\includegraphics[width=0.09\textwidth]{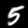}\!\!\! &
\!\!\!\includegraphics[width=0.09\textwidth]{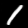}\!\!\! &
\!\!\!\includegraphics[width=0.09\textwidth]{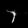}\!\!\! &
\!\!\!\includegraphics[width=0.09\textwidth]{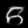}\!\!\! &
\!\!\!\includegraphics[width=0.09\textwidth]{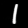}\!\! &
\multicolumn{1}{|c}{\!\!\includegraphics[width=0.09\textwidth]{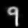}\!\!\!} &
\!\!\!\includegraphics[width=0.09\textwidth]{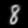}\!\! \\
  \!\!\includegraphics[width=0.09\textwidth]{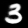}\!\!\! &
\!\!\!\includegraphics[width=0.09\textwidth]{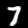}\!\!\! &
\!\!\!\includegraphics[width=0.09\textwidth]{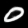}\!\!\! &
\!\!\!\includegraphics[width=0.09\textwidth]{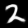}\!\!\! &
\!\!\!\includegraphics[width=0.09\textwidth]{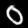}\!\!\! &
\!\!\!\includegraphics[width=0.09\textwidth]{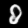}\!\!\! &
\!\!\!\includegraphics[width=0.09\textwidth]{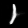}\!\!\! &
\!\!\!\includegraphics[width=0.09\textwidth]{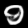}\!\! &
\multicolumn{1}{|c}{\!\!\includegraphics[width=0.09\textwidth]{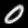}\!\!\!} &
\!\!\!\includegraphics[width=0.09\textwidth]{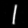}\!\! \\
 $p1$ & $p2$ & $p3$ & $p4$ & $p5$ & $p6$ & $p7$ & $p8$ &\multicolumn{1}{|c}{$p1$} & $p2$ \\[0.1cm]
\multicolumn{8}{|l}{$\qquad\qquad\qquad$(c) $\,d=8$} & \multicolumn{2}{|c|}{$\!$(d)$\,d\!=\!2$} \\[0.1cm]
\hline
\end{tabular}
\caption{Paired eigen-digits
for deep principal components
of Eccentric Autoencoders trained
on MNIST
with latent dimension 6, 4, 8 and 2.}
\label{figure:eigendigits}
\end{figure}
\vspace*{0.2cm}
\end{minipage}

\section{Autoencoder Experiments}
\label{section:auto}

In this section,
we explore the eccentric loss function as a method
for regularizing the latent variables of an
auto\-encoder.
The loss function in this case is:
\[
{\rm loss} = {\rm loss}_{\rm recon} + \lambda\,l_{\mu}(\{\z_i\}),
\]
where ${\rm loss}_{\rm recon}$ is the $L_2$ distance between the
original and reconstructed image, for a training batch $\{\z_i\}$.

\smallskip\noindent
Specifically, we train Eccentric Autoencoders
on:
\begin{enumerate}[label=(\alph*)]
\item MNIST \citep{lecun1998gradient}
for 1000 epochs
using dimension $d=2,4,6,8$
with $\lambda=10^{-\text{\small 3}},\mu=1\,$;
\item CelebA
\citep{liu2015deep}
for 100 epochs
using $\lambda=10^{-\text{\small 4}}, 10^{-\text{\small 3}}$,
\hbox{$\mu=1,1.5,4.5,16.5,64.5$}, with
dimension $d=64\,$.
\end{enumerate}

As far as possible, we try to use the same network structure
and hyperparameters as \citep{tolstikhin2018wasserstein},
with the Adam optimizer \citep{kingma2015adam},
batch size 100, learning rate of $10^{-\text{\small 4}}$
and weight decay of $10^{-\text{\small 6}}$
(see Appendix for details).
Each MNIST run takes approximately 3 hours on a
GeForce GTX 1080 Ti; each CelebA run takes approximately
6 hours on one node of a V100 GPU.

Figure~\ref{figure:spectrum}
shows the eigenvalues for the covariance matrix
of the CelebA
images in latent space, 
for different values of $\lambda$ and $\mu$.
We see that the eigenvalues become more uniform
(closer to $1.0$)
as $\lambda$ increases and, to a lesser extent,
as $\mu$ increases.
When $\lambda$ and $\mu$ decrease, the distribution
becomes more eccentric, with eigenvalues ranging as high
as $3.24$ and as low as $0.47\,$.
As discussed in the previous section,
the sum of the eigenvalues
(indicated by the area under the curve)
is approximately equal to the dimension $d$.
Note that in each case the curves from four independent runs
match each other almost exactly, indicating that the
spectrum is largely invariant from one run to another.

\begin{figure}[t]
\vspace*{-0.2cm}
\hspace*{-0.3cm}\begin{tabular}{cccccccccc}
    \includegraphics[width=0.09\textwidth]{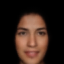}\!\! &
\!\!\includegraphics[width=0.09\textwidth]{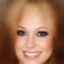}\!\! &
\!\!\includegraphics[width=0.09\textwidth]{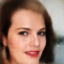}\!\! &
\!\!\includegraphics[width=0.09\textwidth]{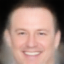}\!\! &
\!\!\includegraphics[width=0.09\textwidth]{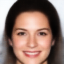}\!\! &
\!\!\includegraphics[width=0.09\textwidth]{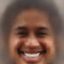}\!\! &
\!\!\includegraphics[width=0.09\textwidth]{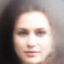}\!\! &
\!\!\includegraphics[width=0.09\textwidth]{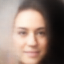}\!\! &
\!\!\includegraphics[width=0.09\textwidth]{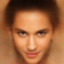}\!\! &
\!\!\includegraphics[width=0.09\textwidth]{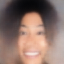} \\
    \includegraphics[width=0.09\textwidth]{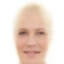}\!\! &
\!\!\includegraphics[width=0.09\textwidth]{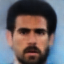}\!\! &
\!\!\includegraphics[width=0.09\textwidth]{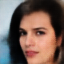}\!\! &
\!\!\includegraphics[width=0.09\textwidth]{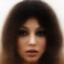}\!\! &
\!\!\includegraphics[width=0.09\textwidth]{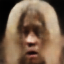}\!\! &
\!\!\includegraphics[width=0.09\textwidth]{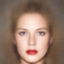}\!\! &
\!\!\includegraphics[width=0.09\textwidth]{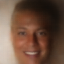}\!\! &
\!\!\includegraphics[width=0.09\textwidth]{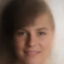}\!\! &
\!\!\includegraphics[width=0.09\textwidth]{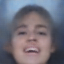}\!\! &
\!\!\includegraphics[width=0.09\textwidth]{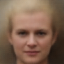} \\
  $p1$ & $p2$ & $p3$ & $p4$ & $p5$ & $p6$ & $p7$ & $p8$ & $p9$ & $p10$ \\[0.1cm]
    \includegraphics[width=0.09\textwidth]{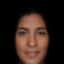}\!\! &
\!\!\includegraphics[width=0.09\textwidth]{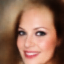}\!\! &
\!\!\includegraphics[width=0.09\textwidth]{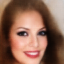}\!\! &
\!\!\includegraphics[width=0.09\textwidth]{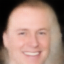}\!\! &
\!\!\includegraphics[width=0.09\textwidth]{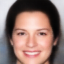}\!\! &
\!\!\includegraphics[width=0.09\textwidth]{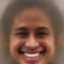}\!\! &
\!\!\includegraphics[width=0.09\textwidth]{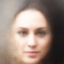}\!\! &
\!\!\includegraphics[width=0.09\textwidth]{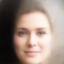}\!\! &
\!\!\includegraphics[width=0.09\textwidth]{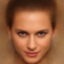}\!\! &
\!\!\includegraphics[width=0.09\textwidth]{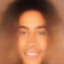} \\
    \includegraphics[width=0.09\textwidth]{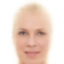}\!\! &
\!\!\includegraphics[width=0.09\textwidth]{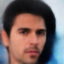}\!\! &
\!\!\includegraphics[width=0.09\textwidth]{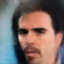}\!\! &
\!\!\includegraphics[width=0.09\textwidth]{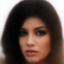}\!\! &
\!\!\includegraphics[width=0.09\textwidth]{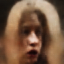}\!\! &
\!\!\includegraphics[width=0.09\textwidth]{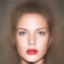}\!\! &
\!\!\includegraphics[width=0.09\textwidth]{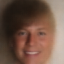}\!\! &
\!\!\includegraphics[width=0.09\textwidth]{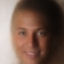}\!\! &
\!\!\includegraphics[width=0.09\textwidth]{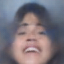}\!\! &
\!\!\includegraphics[width=0.09\textwidth]{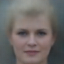} \\
%   $p1$ & $p2$ & $p3$ & $p4$ & $p5$ & $p6$ & $p7$ & $p8$ & $p9$ & $p10$ \\[0.1cm]
%    \includegraphics[width=0.09\textwidth]{eig/celeb7w_pc01_b.png}\!\! &
%\!\!\includegraphics[width=0.09\textwidth]{eig/celeb7w_pc02_b.png}\!\! &
%\!\!\includegraphics[width=0.09\textwidth]{eig/celeb7w_pc03_b.png}\!\! &
%\!\!\includegraphics[width=0.09\textwidth]{eig/celeb7w_pc04_b.png}\!\! &
%\!\!\includegraphics[width=0.09\textwidth]{eig/celeb7w_pc05_a.png}\!\! &
%\!\!\includegraphics[width=0.09\textwidth]{eig/celeb7w_pc06_a.png}\!\! &
%\!\!\includegraphics[width=0.09\textwidth]{eig/celeb7w_pc07_a.png}\!\! &
%\!\!\includegraphics[width=0.09\textwidth]{eig/celeb7w_pc08_a.png}\!\! &
%\!\!\includegraphics[width=0.09\textwidth]{eig/celeb7w_pc09_a.png}\!\! &
%\!\!\includegraphics[width=0.09\textwidth]{eig/celeb7w_pc10_a.png} \\
%    \includegraphics[width=0.09\textwidth]{eig/celeb7w_pc01_a.png}\!\! &
%\!\!\includegraphics[width=0.09\textwidth]{eig/celeb7w_pc02_a.png}\!\! &
%\!\!\includegraphics[width=0.09\textwidth]{eig/celeb7w_pc03_a.png}\!\! &
%\!\!\includegraphics[width=0.09\textwidth]{eig/celeb7w_pc04_a.png}\!\! &
%\!\!\includegraphics[width=0.09\textwidth]{eig/celeb7w_pc05_b.png}\!\! &
%\!\!\includegraphics[width=0.09\textwidth]{eig/celeb7w_pc06_b.png}\!\! &
%\!\!\includegraphics[width=0.09\textwidth]{eig/celeb7w_pc07_b.png}\!\! &
%\!\!\includegraphics[width=0.09\textwidth]{eig/celeb7w_pc08_b.png}\!\! &
%\!\!\includegraphics[width=0.09\textwidth]{eig/celeb7w_pc09_b.png}\!\! &
%\!\!\includegraphics[width=0.09\textwidth]{eig/celeb7w_pc10_b.png} \\
   $p1$ & $p2$ & $p3$ & $p4$ & $p5$ & $p6$ & $p7$ & $p8$ & $p9$ & $p10$ \\[-0.2cm]
\end{tabular}
\caption{Paired eigen-faces corresponding to the first ten deep principal components for
two Eccentric Autoencoders trained independently on CelebA from different random
initial weights.}
%The result is very similar,
%although we can see some intermixing between neighboring components with close eigenvalues,
%such as ($p2,p3$), ($p7,p8$) and ($p9,p10$).}
\label{figure:eigenfaces}
\end{figure}

\begin{figure}[H]
\begin{center}
\vspace*{-0.1cm}
\begin{tabular}{cccccccc}
$\!\!\!$\includegraphics[width=0.09\textwidth]{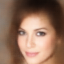}$\!\!\!$ &
$\!\!\!$\includegraphics[width=0.09\textwidth]{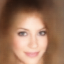}$\!\!\!$ &
  $\!\!$\includegraphics[width=0.09\textwidth]{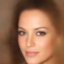}$\!\!\!$ &
$\!\!\!$\includegraphics[width=0.09\textwidth]{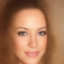}$\!\!$ &
  $\!\!$\includegraphics[width=0.09\textwidth]{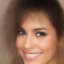}$\!\!\!$ &
$\!\!\!$\includegraphics[width=0.09\textwidth]{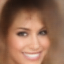}$\!\!\!$ &
  $\!\!$\includegraphics[width=0.09\textwidth]{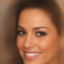}$\!\!\!$ &
$\!\!\!$\includegraphics[width=0.09\textwidth]{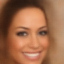}$\!\!\!$ \\[-0.04cm]
$\!\!\!$\includegraphics[width=0.09\textwidth]{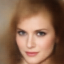}$\!\!\!$ &
$\!\!\!$\includegraphics[width=0.09\textwidth]{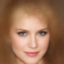}$\!\!\!$ &
  $\!\!$\includegraphics[width=0.09\textwidth]{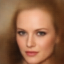}$\!\!\!$ &
$\!\!\!$\includegraphics[width=0.09\textwidth]{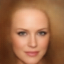}$\!\!$ &
  $\!\!$\includegraphics[width=0.09\textwidth]{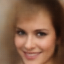}$\!\!\!$ &
$\!\!\!$\includegraphics[width=0.09\textwidth]{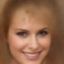}$\!\!\!$ &
  $\!\!$\includegraphics[width=0.09\textwidth]{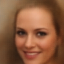}$\!\!\!$ &
$\!\!\!$\includegraphics[width=0.09\textwidth]{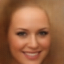}$\!\!\!$ \\
$\!\!\!$\includegraphics[width=0.09\textwidth]{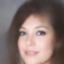}$\!\!\!$ &
$\!\!\!$\includegraphics[width=0.09\textwidth]{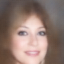}$\!\!\!$ &
  $\!\!$\includegraphics[width=0.09\textwidth]{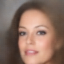}$\!\!\!$ &
$\!\!\!$\includegraphics[width=0.09\textwidth]{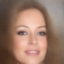}$\!\!$ &
  $\!\!$\includegraphics[width=0.09\textwidth]{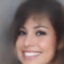}$\!\!\!$ &
$\!\!\!$\includegraphics[width=0.09\textwidth]{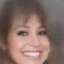}$\!\!\!$ &
  $\!\!$\includegraphics[width=0.09\textwidth]{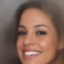}$\!\!\!$ &
$\!\!\!$\includegraphics[width=0.09\textwidth]{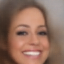}$\!\!\!$ \\[-0.04cm]
$\!\!\!$\includegraphics[width=0.09\textwidth]{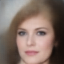}$\!\!\!$ &
$\!\!\!$\includegraphics[width=0.09\textwidth]{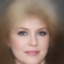}$\!\!\!$ &
  $\!\!$\includegraphics[width=0.09\textwidth]{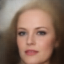}$\!\!\!$ &
$\!\!\!$\includegraphics[width=0.09\textwidth]{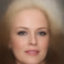}$\!\!$ &
  $\!\!$\includegraphics[width=0.09\textwidth]{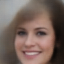}$\!\!\!$ &
$\!\!\!$\includegraphics[width=0.09\textwidth]{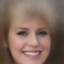}$\!\!\!$ &
  $\!\!$\includegraphics[width=0.09\textwidth]{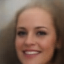}$\!\!\!$ &
$\!\!\!$\includegraphics[width=0.09\textwidth]{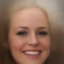}$\!\!\!$ \\[0.04cm]
$\!\!\!$\includegraphics[width=0.09\textwidth]{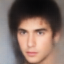}$\!\!\!$ &
$\!\!\!$\includegraphics[width=0.09\textwidth]{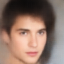}$\!\!\!$ &
  $\!\!$\includegraphics[width=0.09\textwidth]{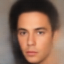}$\!\!\!$ &
$\!\!\!$\includegraphics[width=0.09\textwidth]{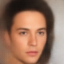}$\!\!$ &
  $\!\!$\includegraphics[width=0.09\textwidth]{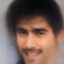}$\!\!\!$ &
$\!\!\!$\includegraphics[width=0.09\textwidth]{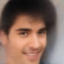}$\!\!\!$ &
  $\!\!$\includegraphics[width=0.09\textwidth]{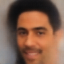}$\!\!\!$ &
$\!\!\!$\includegraphics[width=0.09\textwidth]{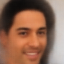}$\!\!\!$ \\[-0.04cm]
$\!\!\!$\includegraphics[width=0.09\textwidth]{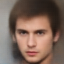}$\!\!\!$ &
$\!\!\!$\includegraphics[width=0.09\textwidth]{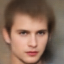}$\!\!\!$ &
  $\!\!$\includegraphics[width=0.09\textwidth]{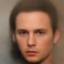}$\!\!\!$ &
$\!\!\!$\includegraphics[width=0.09\textwidth]{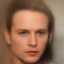}$\!\!$ &
  $\!\!$\includegraphics[width=0.09\textwidth]{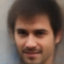}$\!\!\!$ &
$\!\!\!$\includegraphics[width=0.09\textwidth]{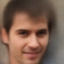}$\!\!\!$ &
  $\!\!$\includegraphics[width=0.09\textwidth]{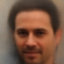}$\!\!\!$ &
$\!\!\!$\includegraphics[width=0.09\textwidth]{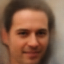}$\!\!\!$ \\
$\!\!\!$\includegraphics[width=0.09\textwidth]{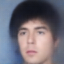}$\!\!\!$ &
$\!\!\!$\includegraphics[width=0.09\textwidth]{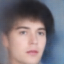}$\!\!\!$ &
  $\!\!$\includegraphics[width=0.09\textwidth]{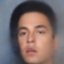}$\!\!\!$ &
$\!\!\!$\includegraphics[width=0.09\textwidth]{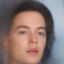}$\!\!$ &
  $\!\!$\includegraphics[width=0.09\textwidth]{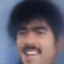}$\!\!\!$ &
$\!\!\!$\includegraphics[width=0.09\textwidth]{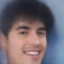}$\!\!\!$ &
  $\!\!$\includegraphics[width=0.09\textwidth]{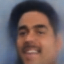}$\!\!\!$ &
$\!\!\!$\includegraphics[width=0.09\textwidth]{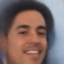}$\!\!\!$ \\[-0.04cm]
$\!\!\!$\includegraphics[width=0.09\textwidth]{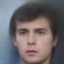}$\!\!\!$ &
$\!\!\!$\includegraphics[width=0.09\textwidth]{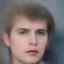}$\!\!\!$ &
  $\!\!$\includegraphics[width=0.09\textwidth]{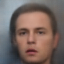}$\!\!\!$ &
$\!\!\!$\includegraphics[width=0.09\textwidth]{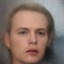}$\!\!$ &
  $\!\!$\includegraphics[width=0.09\textwidth]{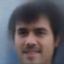}$\!\!\!$ &
$\!\!\!$\includegraphics[width=0.09\textwidth]{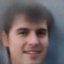}$\!\!\!$ &
  $\!\!$\includegraphics[width=0.09\textwidth]{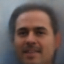}$\!\!\!$ &
$\!\!\!$\includegraphics[width=0.09\textwidth]{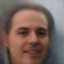}$\!\!\!$\\[-0.2cm]
\end{tabular}
\caption{Images generated from principal components
$p6$, $p11$, $p3$ (horizontally) and $p2$, $p10$, $p9$ (vertically).}
\label{figure:tess}
\end{center}
\end{figure}

\begin{figure}[H]
\hspace*{-0.2cm}\begin{tabular}{cccc}
$\!$\includegraphics[width=0.24\textwidth]{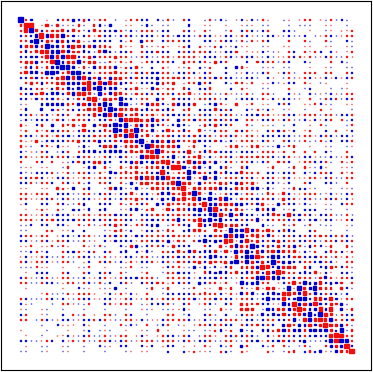}$\!\!$ &
$\!\!$\includegraphics[width=0.24\textwidth]{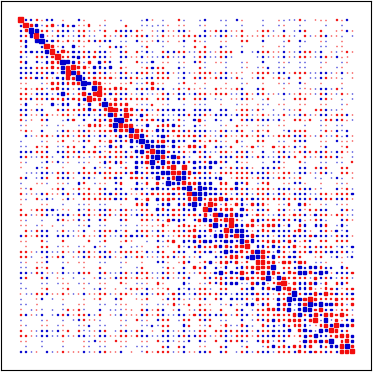}$\!$ &
$\!$\includegraphics[width=0.24\textwidth]{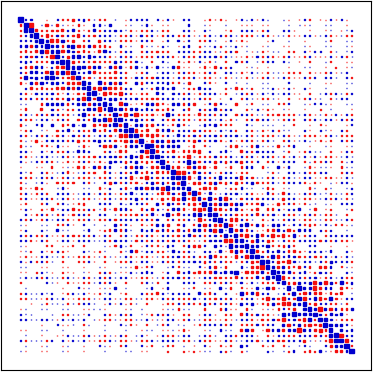}$\!\!$ &
$\!\!$\includegraphics[width=0.24\textwidth]{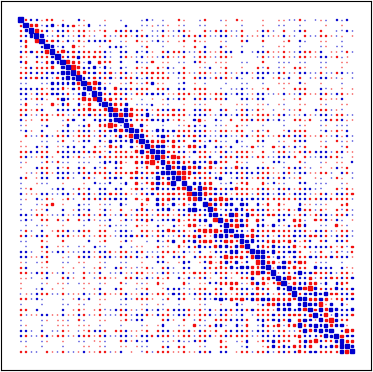}$\!$ \\
   (a) $\lambda=10^{-\text{\small 3}}$, raw & (b) $\lambda=10^{-\text{\small 4}}$, raw &
  (c) $\lambda=10^{-\text{\small 3}}$, aligned & (d) $\lambda=10^{-\text{\small 4}}$, aligned \\
\end{tabular}
\caption{Cross correlation between deep principal components
for Eccentric Autoencoders
%with $\mu=1$ and $\lambda=0.001$ or $0.0001$,
trained independently~on the CelebA dataset
from different random initial weights,
before the alignment procedure and afterwards.}
\label{figure:correlation}
\end{figure}

\subsection{Principal Components and Visualization}

If we shift and rotate the latent vectors of the training items
so that their mean becomes zero and their coordinate axes lie
along the eigenvectors of the covariance matrix
in decreasing order,
we can extract pairs of deep eigen-digits or eigen-faces
by choosing (appropriately scaled) positive and negative
basis vectors along each of these principal axes.
Eigen-digits for MNIST with different latent dimension
are shown in Figure~\ref{figure:eigendigits},
and eigen-faces for CelebA
from two different runs
are shown in
Figure \ref{figure:eigenfaces}.
The principal components from the two runs are very similar,
although we can see some intermixing between neighboring components with close eigenvalues,
such as ($p2,p3$), ($p7,p8$) and ($p9,p10$).

This decomposition could potentially serve as the basis for a
visualization tool
or an
interactive interface for generating desired images.
Suppose, for example,
that we wish to further explore
the effect of components $p2, p3, p6, p9, p10$ and $p11$.
%in Figure~\ref{figure:eigenfaces} (top row).
We could generate
the plot shown in Figure~\ref{figure:tess} by
making the selected coordinates positive and negative in all possible combinations.

When contemplating such a visualization tool,
we believe there are certain advantages to
be gained from allowing unevenness
or eccentricity
in the eigenvalues of the covariance matrix.
For example, it may concentrate the intrinsic variation
more heavily into the first few principal components,
thus reducing the cognitive load for the human user
and alerting them to the relative prominence of
the various latent factors.
In addition, it may provide a greater degree of
consistency from
one run to another, allowing a relatively
stable set of principal components to be extracted
when the algorithm is re-run with different initial weights.

\subsection{Representation Alignment and Visualization}

\smallskip
The full mapping to the rotated latent space can be
considered
as an encoding which maps each input image
to a vector $(p_1,\ldots,p_d)$ where $p_k$
is the value of the coordinate in the direction
of the $k$th (deep) principal component.
We are interested in the question: How canonical
is this encoding? In other words, if we train
two autoencoders independently from different
random weights, producing two encodings
$E_1: {\rm Img}\mapsto (p_1,\ldots,p_d)$ and
$E_2: {\rm Img}\mapsto (q_1,\ldots,q_d)$,
how similar are these two mappings?

The images on the left side of
Figure~\ref{figure:correlation} illustrate the cor\-relation
between the principal components of two such mappings.
We note that it is possible for
principal components
corresponding to close eigenvalues to be permuted,
for example $p2$ and $p3$ in
Figure~\ref{figure:eigendigits} (a) and (c),
which would appear as slightly off-diagonal terms in
Figure~\ref{figure:correlation}.
It is also possible for the positive and negative
eigenvectors to be inverted, for example $p4$ in
Figure~\ref{figure:eigendigits} (a) and (c),
which would appear in Figure~\ref{figure:correlation}
as diagonal terms which are negative (red) rather than positive (blue).

\begin{table}[t]
\begin{center}
\vspace*{-0.2cm}
%{\small
\begin{tabular}{lcccccc}
\toprule
$\!\!\!\!\!$Algorithm$\!\!\!\!\!\!\!\!$ & $\lambda$ & $\!\!\!\!\!\!\varepsilon\!\!$ & $\!\!\!\!$angle(raw)$\!\!\!$ & $\!\!\!$angle(align)$\!\!\!\!$ & FI\ score \\[0.09cm]
\midrule
EAE & $10^{-\text{\small 2}}$ & 0.092 & 81$^\circ \rpm$ 1$^\circ$ & 69$^\circ \rpm$ 1$^\circ$ & 53.7 $\rpm$ 1.2 \\
EAE & $10^{-\text{\small 3}}$ & 0.29 & 69$^\circ \rpm$ 2$^\circ$ & 54$^\circ \rpm$ 1$^\circ$ & 50.2 $\rpm$ 0.5 \\
EAE & $\!\!\!\!3\!\times\!10^{-\text{\small 4}}$ & 0.49 & 59$^\circ \rpm$ 1$^\circ$ & 47$^\circ \rpm$ 1$^\circ$ & 48.7 $\rpm$ 1.0 \\
EAE & $10^{-\text{\small 4}}$ & 0.54 & 53$^\circ \rpm$ 2$^\circ$ & 43$^\circ \rpm$ 1$^\circ$ & 49.5 $\rpm$ 0.9 \\
EAE & $10^{-\text{\small 5}}$ & 0.71 & 55$^\circ \rpm$ 5$^\circ$ & 39$^\circ \rpm$ 1$^\circ$ & 50.3 $\rpm$ 0.8 \\
EAE & $10^{-\text{\small 6}}$ & 0.75 & 50$^\circ \rpm$ 2$^\circ$ & 40$^\circ \rpm$ 1$^\circ$ & 55.0 $\rpm$ 1.7 \\[0.09cm]
\hline \\[-0.2cm]
%VAE & $10^{-\text{\small 1}}$ & & $\!\!$19.55 $\rpm$ 0.08 & 28$^\circ \rpm$ 75$^\circ$ & 22$^\circ \rpm$ 4$^\circ$ \\[0.1cm]
%VAE & $10^{-\text{\small 2}}$ & &  7.99 $\rpm$ 0.01 & 57$^\circ \rpm$ 3$^\circ$ & 43$^\circ \rpm$ 2$^\circ$\\
VAE & $10^{-\text{\small 3}}$ &   0.25 & 76$^\circ \rpm$ 2$^\circ$ & 63$^\circ \rpm$ 1$^\circ$ & 50.0 $\rpm$ 0.7 \\
VAE & $\!\!\!\!3\!\times\!10^{-\text{\small 4}}$ & 0.41 & 71$^\circ \rpm$ 2$^\circ$ & 58$^\circ \rpm$ 1$^\circ$ & 53.7 $\rpm$ 1.9 \\
VAE & $10^{-\text{\small 4}}$ &   0.56 & 59$^\circ \rpm$ 1$^\circ$ & 46$^\circ \rpm$ 1$^\circ$ & 57.7 $\rpm$ 1.0 \\
VAE & $10^{-\text{\small 5}}$ &   0.71 & 51$^\circ \rpm$ 2$^\circ$ & 40$^\circ \rpm$ 1$^\circ$ & 57.8 $\rpm$ 0.2 \\[0.1cm]
\hline \\[-0.2cm]
WAE & $10^{-\text{\small 1}}$ &  0.19 & 79$^\circ \rpm$ 1$^\circ$ & 65$^\circ \rpm$ 1$^\circ$ & 53.5 $\rpm$ 0.8 \\
WAE & $10^{-\text{\small 2}}$ &  0.29 & 62$^\circ \rpm$ 2$^\circ$ & 49$^\circ \rpm$ 1$^\circ$ & 49.2 $\rpm$ 1.4 \\
WAE & $\!\!\!\!3\!\times\!10^{-\text{\small 3}}$ & 0.45 & 53$^\circ \rpm$ 4$^\circ$ & 40$^\circ \rpm$ 1$^\circ$ & 49.4 $\rpm$ 0.6 \\
WAE & $10^{-\text{\small 3}}$ &  0.59 & 51$^\circ \rpm$ 3$^\circ$ & 40$^\circ \rpm$ 2$^\circ$ & 50.0 $\rpm$ 1.0 \\
WAE & $\!\!\!\!3\!\times\!10^{-\text{\small 4}}$ & 0.69 & 52$^\circ \rpm$ 2$^\circ$ & 40$^\circ \rpm$ 1$^\circ$ & 51.2 $\rpm$ 0.6 \\
WAE & $10^{-\text{\small 4}}$ &  0.76 & 53$^\circ \rpm$ 5$^\circ$ & 41$^\circ \rpm$ 1$^\circ$ & 54.2 $\rpm$ 0.4 \\[0.1cm]
\hline \\[-0.2cm]
AE & $0$ & 0.71 & 48$^\circ \rpm$ 1$^\circ$ & 38$^\circ \rpm$ 1$^\circ$ & 55.8 $\rpm$ 0.1 \\
\bottomrule
\end{tabular}
%}
\end{center}
\caption{Eccentricity ($\varepsilon$)
of distribution in latent space,
mean angle (raw and aligned) between 
independently trained principal component encodings,
and Fr\'echet Inception (FI) score,
for Autoencoder models trained on Celeba
with different values of the scaling parameter $\lambda$.}
%using EAE, VAE and WAE.}
%and Wasserstein Autoencoder (WAE).}
\label{table:frechet}
\end{table}

In order to resolve these ambiguities
and try to bring the two embeddings into
approximate alignment,
we use the Hungarian Algorithm
\citep{kuhn1955hungarian}
to find a permutation which maximizes the trace
of the absolute
(training set) cross-covariance between the two mappings,
and then flip the signs of individual components
to make the cross-covariance matrix positive along the diagonal.
This alignment procedure can be visualized by comparing the
(test set) correlations on the left of Figure~\ref{figure:correlation}
with those on the right.
It has the effect of
bringing the correlation closer to the diagonal, and turning it from
negative (red) to positive (blue).  The correlation appears to
be more heavily concentrated along the diagonal for
$\lambda=10^{-\text{\small 4}}$ than for
$\lambda=10^{-\text{\small 3}}$.
The appearance of a
\hbox{2-by-2} block along the diagonal
with three blue squares and one red square,
such as in the top left corner of
Figure~\ref{figure:correlation}(d),
is indicative of a rotation in the
\hbox{2-dimensional} subspace formed by these two components.
The alignment between representations from different runs
can be quantified by measuring the average angle between
corresponding vectors in the two representations
(shown in Table~1, and discussed in the next subsection).

\subsection{Eccentricity, Alignment and Fr\'echet Inception Score}

Trained autoencoders can be used for image generation,
with latent vectors chosen randomly from
either a standard normal or
multivariate Gaussian distribution.
The Fr\'echet Inception (FI) score
has become a standard tool for measuring how well
the distribution of generated images conforms to
that of an unseen set of test images.
In the case where the latent vectors
are chosen from a standard normal distribution,
\cite{tolstikhin2018wasserstein}
report an FI score of 63 for VAE,
55 for WAE-MMD and 42 for WAE-GAN
(which includes a discriminator).
In this section,
we use a scaling parameter to modulate the eccentricity
of the distribution,
and explore image generation
using latent vectors chosen from a
multivariate Gaussian whose mean and covariance match
those of the encoded latent vectors from the training images.

\begin{figure}[t]
\hspace*{1cm}
\includegraphics[width=0.8\textwidth]{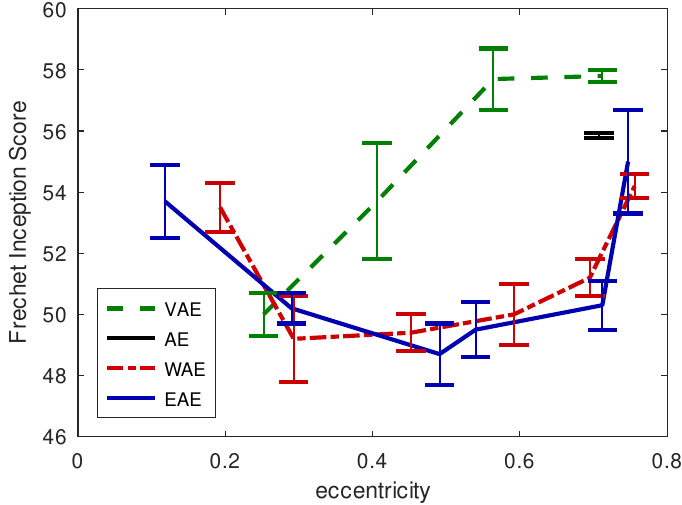}
\caption{Relationship between eccentricity ($\varepsilon$) and
Fr\'echet Inception (FI) score,
for models listed in Table~\ref{table:frechet}.}
\label{figure:frechet}
\end{figure}

Table~\ref{table:frechet}
shows the mean angle between encodings of the same
(test set) images,
averaged across six pairwise combinations of
four runs of our Eccentric Autoencoder (EAE),
for different choices of $\lambda$,
along with the Fr\'echet Inception score
and the eccentricity, which we define to be
the standard deviation of the
eigenvalues of the covariance matrix
divided by their mean.
For comparison, we include VAE \citep{DBLP:journals/corr/KingmaW13}
and WAE-MMD \citep{tolstikhin2018wasserstein},
as well as a plain autoencoder (AE) with no regularization
other than weight decay.

Each of the three regularizers shows a similar pattern:
When $\lambda$ is (relatively) large,
the distribution in latent space becomes close to
both the spherical and the standard normal distribution.
As $\lambda$ decreases, the distribution becomes less constrained
and gradually closer to that of an unregulated autoencoder (AE),
which has an eccentricity
$\varepsilon$ of 0.71,
and generates images with an FI score of 55.8.

Our Eccentric Autoencoder (EAE)
and the Wasserstein Autoencoder (WAE)
both achieve a minimum FI score of around 49 or 50
for $\varepsilon$ in the range of 0.3 to 0.7
(see Figure~\ref{figure:frechet}).
The FI score increases to around 54 when
$\varepsilon$~is~low (over-regulation)
and when it is high (under-regulation).
For the Variational Autoencoder (VAE)
the FI score is around 50 for $\varepsilon\simeq$ 0.25 
but increases to 58 when $\varepsilon\simeq$ 0.56.
It should be noted that here we are only comparing different
methods for regularizing autoencoders,
and that considerably better
image generation can now be achieved by
other methods such as
2-Stage VAE \citep{dai2019diagnosing},
BigGAN \citep{brock2018large}
or diffusion models \citep{dhariwal2021diffusion}.

As explained in the previous subsection,
modulating for increased eccentricity
may have the additional benefit of
allowing deep principal components
to be extracted in a manner that is
relatively invariant from one run to another.
This is quantified in the 5th column of Table~1,
where the mean angle between corresponding
latent vectors across different runs reaches a
minimum of 39 or 40 degrees for $\varepsilon$
greater than about 0.45 (WAE) or 0.6 (EAE).

\subsection{Downstream Classification}

Another way to test the efficacy of an autoencoder
is to train a classifier on the latent variables
using different sized subsets of labeled training items
and measure the performance.
Table~\ref{table:semisupervised} shows the result
of training a KNN classifier on subsets
of labeled training items of the specified size,
averaged across three autoencoder runs
and 20 random subsets for each run.
For comparison, we note that
the results for $d=8$ and training size $\ge 100$
are very close (in fact, within the margin of error)
to those reported for IRMAE in
\citep{jing2020implicit} although in that case
an MLP with two fully connected layers of dimension 128
was used for classification rather than KNN.

\begin{table}[H]
\begin{center}
%{\small
\begin{tabular}{crrrrr}
\toprule
  &   10(1)$\;\;$ & 100(1)$\;$ & 1000(5)$\;$ & $\!\!\!\!\!\!$10000(10) & $\!\!\!\!\!\!\!\!$60000(15)$\!\!\!\!$ \\
\midrule
$\!\!d\!=\,$2 & $\!\!\!$39.0 $\!\rpm\!$ 7.2$\!$ & $\!$19.1 $\!\rpm\!$ 2.6$\!$ & $\!\!\!\!\!$12.4 $\!\rpm\!$ 1.3 & $\!\!\!\!$10.9
$\!\rpm\!$ 0.9 & $\!\!\!\!\!\!$10.6 $\!\rpm\!$ 0.8$\!\!$ \\
$\!\!d\!=\,$4 & $\!\!\!$36.0 $\!\rpm\!$ 6.6$\!$ & $\!$12.1 $\!\rpm\!$ 1.9$\!$ & 5.4 $\!\rpm\!$ 0.6 & 4.0 $\!\rpm\!$ 0.3 & 3.7 $\!\rpm\!$ 0.3$\!\!$ \\
$\!\!d\!=\,$6 & $\!\!\!$33.7 $\!\rpm\!$ 7.3$\!$ & $\!$10.0 $\!\rpm\!$ 1.3$\!$ & 4.1 $\!\rpm\!$ 0.3 & 3.0 $\!\rpm\!$ 0.1 & 2.6 $\!\rpm\!$ 0.1$\!\!$ \\
$\!\!d\!=\,$8 & $\!\!\!$34.0 $\!\rpm\!$ 5.1$\!$ & 9.5 $\!\rpm\!$ 1.1$\!$ & 3.7 $\!\rpm\!$ 0.2 &
2.4 $\!\rpm\!$ 0.1 & 2.0 $\!\rpm\!$ 0.1$\!\!$ \\
\bottomrule
\end{tabular}
%}
\end{center}
\caption{Error Rate for Downstream Classification on MNIST
with different numbers of labeled training items,
using a KNN classifier (number of neighbors
specified in parentheses).}
\label{table:semisupervised}
\end{table}

\begin{table}[H]
\begin{center}
%{\small
\begin{tabular}{crccc}
\toprule
$\lambda$ & $\mu\;$ & DT(1) & DT(8) & KNN(20) \\
\midrule
$10^{-\text{\small 4}}$ &  1.0 & $\!$16.68 $\rpm$ 0.12$\!$ & $\!$14.42 $\rpm$ 0.11$\!$ & $\!$13.15 $\rpm$ 0.03$\!$ \\
$10^{-\text{\small 4}}$ & $\!\!$64.5 & $\!$17.65 $\rpm$ 0.33$\!$ & $\!$14.56 $\rpm$ 0.16$\!$ & $\!$13.14 $\rpm$ 0.04$\!$ \\
$10^{-\text{\small 3}}$ &  1.0 & $\!$17.61 $\rpm$ 0.37$\!$ & $\!$15.20 $\rpm$ 0.19$\!$ & $\!$13.30 $\rpm$ 0.01$\!$ \\
$10^{-\text{\small 3}}$ & $\!\!$64.5 & $\!$17.90 $\rpm$ 0.32$\!$ & $\!$15.52 $\rpm$ 0.17$\!$ & $\!$13.43 $\rpm$ 0.02$\!$ \\
\bottomrule
\end{tabular}
\end{center}
%}
\caption{Classification error averaged over
40 attributes for CelebA, using
Decision Tree of depth~1 (Stump),
Decision Tree with maximum depth 8,
and KNN with 20 neighbors.}
\label{table:celeba_class}
\end{table}

\begin{table}[t]
\begin{center}
\vspace*{-0.2cm}
%{\small
\begin{tabular}{llc}
\toprule
$\!\!$Attribute & $\!\!\!\!\!\!\!\!$Component & Information Gain \\
\midrule
$\!\!$\phantom{2}3. Attractive & $p2$ & 50\% $\rightarrow$ 65\% \\
$\!\!$19. Heavy Makeup & $p2$ & 60\% $\rightarrow$ 76\% \\
$\!\!$20. High Cheek Bones & $p2$ & 52\% $\rightarrow$ 68\% \\
$\!\!$21. Male & $p2$ & 61\%  $\rightarrow$ 80\% \\
$\!\!$22. Mouth Slightly Open & $p6$ & 50\% $\rightarrow$ 66\% \\
$\!\!$32. Smiling & $p6$ & 50\% $\rightarrow$ 67\% \\
$\!\!$37. Wearing Lipstick & $p2$ & 52\% $\rightarrow$ 79\% \\
\bottomrule
\end{tabular}
%}
\end{center}
\caption{Attributes for which a single
principal component \\
provides
a~significant information gain.}
\label{table:stump}
\end{table}

Table~\ref{table:celeba_class} shows the
classification error averaged across 40 standard
binary features for CelebA, using a Decision Tree
with depth 1 (also known as a Stump),
a Decision Tree with depth 8, and a KNN
with 20 neighbors. As a baseline,
choosing the majority class in each case would
achieve an error rate of 20.0\%.
Attributes for which a single component
provides a significant information gain are listed
in Table~\ref{table:stump}, with reference to the
principal components shown in
Figure~\ref{figure:eigenfaces}.
According to Table~\ref{table:stump},
the images in the upper half of
Figure~\ref{figure:tess} should appear
more attractive, more female, with higher cheek bones,
wearing lipstick and heavier makeup,
compared to those in the lower half.
Those in the right half should be smiling with
mouth slightly open, compared to those on the left.

\section{Conclusion}

We have introduced the Eccentric Loss function
and shown that it reaches its minimum on a hyperspherical distribution
in dimension $d$
with radius very close to $\sqrt{d}$.

By adjusting the scaling factor, we can force the items to
adhere closely to the hypersphere, or we can enable more
flexibility and tolerance of eccentricity, thus allowing
the latent factors to be stratified according to their
relative importance,
with potential benefits for data visualization and analysis.

Although we have demonstrated our method on image data,
it could in principle be applied to
other kinds of dataset, with an appropriate choice of generator
and encoder network.
In order to avoid explicit tuning, the scaling
factor $\lambda$ could instead be regulated dynamically to achieve
a certain pre-determined target value for the
loss function, or the eccentricity.

In future work, we plan to
explore the use of eccentric regularization
to encourage diversity of latent features in
deep classification networks and BiGANs,
or as a stabilizing component for
recurrent network architectures.

% jjj

%   80.039   80.039   80.039   80.039   80.039

%  DT(1) for CelebA
%   81.188   81.347   81.316   81.470   81.443
%   82.265   81.961   81.978   81.204   81.639

%  DT(10) classification for CelebA
%   84.837   84.599   84.450   84.631   84.428
%   85.670   85.760   85.761   85.381   85.672

% attributes for which DT(1) improves:
%   2,18,19,20,21,31,36,
%   which PC is chosen?

%  knn(20)classification for CelebA
%  Series 1:
%   86.743   86.635   86.612   86.631   86.631
%   86.905   86.921   86.902   86.880   86.940
%  Series 2:
%   86.743   86.668   86.645   86.569   86.569
%   86.837   86.830   86.910   86.848   86.863

%   86.74(0.01) 86.65(0.02) 86.63(0.02) 86.60(0.03) 86.60(0.03)
%   86.87(0.03) 86.88(0.05) 86.91(.001) 86.86(0.02) 86.90(0.04)

%
% l = 0.001,  mu=1     86.74(0.01)
% l = 0.001,  mu=64.5  86.60(0.03)
% l = 0.0001, mu=1     86.87(0.03)
% l = 0.0001, mu=64.5  86.90(0.04)
%

% cross CelebA:

% [old] 8j7r, 8n7v, 8d7w, 8i8a

% 8r7r or 8q7r 7w8b
% (8q7r has rotation)

\bibliography{plum_arxiv}

\begin{thebibliography}{}

\bibitem[Bao et~al., 2013]{bao2013incoherent}
Bao, Y., Jiang, H., Dai, L., and Liu, C. (2013).
\newblock Incoherent training of deep neural networks to de-correlate
  bottleneck features for speech recognition.
\newblock In {\em ICASSP}, pages 6980--6984.

\bibitem[Brock et~al., 2019]{brock2018large}
Brock, A., Donahue, J., and Simonyan, K. (2019).
\newblock Large scale {GAN} training for high fidelity natural image synthesis.
\newblock In {\em ICLR}.

\bibitem[Dai and Wipf, 2019]{dai2019diagnosing}
Dai, B. and Wipf, D. (2019).
\newblock Diagnosing and enhancing {VAE} models.
\newblock {\em arXiv preprint arXiv:1903.05789}.

\bibitem[Davidson et~al., 2018]{DBLP:conf/uai/DavidsonFCKT18}
Davidson, T.~R., Falorsi, L., Cao, N.~D., Kipf, T., and Tomczak, J.~M. (2018).
\newblock Hyperspherical variational auto-encoders.
\newblock In Globerson, A. and Silva, R., editors, {\em UAI}, pages 856--865.

\bibitem[Dhariwal and Nichol, 2021]{dhariwal2021diffusion}
Dhariwal, P. and Nichol, A. (2021).
\newblock Diffusion models beat {GAN}s on image synthesis.
\newblock In {\em NeurIPS}, pages 8780--8794.

\bibitem[Ghosh et~al., 2020]{DBLP:conf/iclr/GhoshSVBS20}
Ghosh, P., Sajjadi, M. S.~M., Vergari, A., Black, M.~J., and Sch{\"{o}}lkopf,
  B. (2020).
\newblock From variational to deterministic autoencoders.
\newblock In {\em ICLR}.

\bibitem[Grover and Ermon, 2019]{grover2019uncertainty}
Grover, A. and Ermon, S. (2019).
\newblock Uncertainty autoencoders: Learning compressed representations via
  variational information maximization.
\newblock In {\em AISTATS}, pages 2514--2524.

\bibitem[Higgins et~al., 2017]{DBLP:conf/iclr/HigginsMPBGBML17}
Higgins, I., Matthey, L., Pal, A., Burgess, C., Glorot, X., Botvinick, M.,
  Mohamed, S., and Lerchner, A. (2017).
\newblock beta-{VAE}: Learning basic visual concepts with a constrained
  variational framework.
\newblock In {\em ICLR}.

\bibitem[Jing et~al., 2020]{jing2020implicit}
Jing, L., Zbontar, J., et~al. (2020).
\newblock Implicit rank-minimizing autoencoder.
\newblock In {\em NeurIPS}, pages 14736--46.

\bibitem[Kingma and Ba, 2015]{kingma2015adam}
Kingma, D.~P. and Ba, J. (2015).
\newblock Adam: A method for stochastic optimization.
\newblock In {\em International Conference on Learning Representations
  ({ICLR})}.

\bibitem[Kingma and Welling, 2014]{DBLP:journals/corr/KingmaW13}
Kingma, D.~P. and Welling, M. (2014).
\newblock Auto-encoding variational {B}ayes.
\newblock In {\em International Conference on Learning Representations
  ({ICLR})}.

\bibitem[Kuhn, 1955]{kuhn1955hungarian}
Kuhn, H.~W. (1955).
\newblock The {H}ungarian method for the assignment problem.
\newblock {\em Naval research logistics quarterly}, 2(1-2):83--97.

\bibitem[LeCun et~al., 1998]{lecun1998gradient}
LeCun, Y., Bottou, L., Bengio, Y., and Haffner, P. (1998).
\newblock Gradient-based learning applied to document recognition.
\newblock {\em Proceedings of the IEEE}, 86(11):2278--2324.

\bibitem[Lin et~al., 2020]{lin2020regularizing}
Lin, R., Liu, W., Liu, Z., Feng, C., Yu, Z., Rehg, J.~M., Xiong, L., and Song,
  L. (2020).
\newblock Regularizing neural networks via minimizing hyperspherical energy.
\newblock In {\em CVPR}, pages 6917--6927.

\bibitem[Liu et~al., 2018]{liu2018learning}
Liu, W., Lin, R., Liu, Z., Liu, L., Yu, Z., Dai, B., and Song, L. (2018).
\newblock Learning towards minimum hyperspherical energy.
\newblock In {\em Advances in Neural Information Processing Systems}, pages
  6222--6233.

\bibitem[Liu et~al., 2015]{liu2015deep}
Liu, Z., Luo, P., Wang, X., and Tang, X. (2015).
\newblock Deep learning face attributes in the wild.
\newblock In {\em ICCV}, pages 3730--3738.

\bibitem[Thomson, 1904]{thomson1904xxiv}
Thomson, J.~J. (1904).
\newblock On the structure of the atom: an investigation of the stability and
  periods of oscillation of a number of corpuscles arranged at equal intervals
  around the circumference of a circle; with application of the results to the
  theory of atomic structure.
\newblock {\em The London, Edinburgh, and Dublin Philosophical Magazine and
  Journal of Science}, 7(39):237--265.

\bibitem[Tolstikhin et~al., 2018]{tolstikhin2018wasserstein}
Tolstikhin, I., Bousquet, O., Gelly, S., and Sch{\"o}lkopf, B. (2018).
\newblock Wasserstein auto-encoders.
\newblock In {\em ICLR}.

\bibitem[van~den Oord et~al., 2017]{DBLP:conf/nips/OordVK17}
van~den Oord, A., Vinyals, O., and Kavukcuoglu, K. (2017).
\newblock Neural discrete representation learning.
\newblock In {\em NeurIPS}, pages 6306--6315.

\bibitem[Xie et~al., 2015]{xie2015diversifying}
Xie, P., Deng, Y., and Xing, E. (2015).
\newblock Diversifying restricted {B}oltzmann machine for document modeling.
\newblock In {\em Proceedings of the 21th ACM SIGKDD International Conference
  on Knowledge Discovery and Data Mining}, pages 1315--1324.

\bibitem[Xie et~al., 2017]{xie2017uncorrelation}
Xie, P., Singh, A., and Xing, E.~P. (2017).
\newblock Uncorrelation and evenness: a new diversity-promoting regularizer.
\newblock In {\em ICML}, pages 3811--3820.

\bibitem[Yu et~al., 2011]{yu2011diversity}
Yu, Y., Li, Y.-F., and Zhou, Z.-H. (2011).
\newblock Diversity regularized machine.
\newblock In {\em Int'l Joint Conference on Artificial Intelligence (IJCAI)},
  pages 1603--1608.

\end{thebibliography}
\bibliographystyle{apalike}

\newpage

\appendix
\section{Appendix}

\subsection{\bf Proof of Theorem 1}

The Spherical Distribution $\Srho$
will be a stationary point % (minimum)
provided the integral, over ${\cal S}_\rho$,
of the gradient of $K(\z_0,\z_\rho)$
is equal to zero for any arbitrary point
$\z_0$ on the sphere, i.e.
\begin{align}
\int_{\z_\rho\in\Sphere_\rho}\nabla_{\z_0}K(\z_0,\z_\rho)\,d\z_\rho &= 0.\qquad\qquad \nonumber \\
%\frac{1}{2d+1}
\hbox{Equivalently},\qquad\qquad
\int_{\Sphere_\rho}
\frac{2\mu(\z_0-\z_\rho)}{1+\frac{||\z_0-\z_\rho||^2}{\N}}\,d\z_\rho
&=\;\z_0. \nonumber
\end{align}

%It will be convenient to split
%$\z = (z_1,z_2,\ldots,z_d)$
%into its first component $\z_1$
%and the remaining components
%$\hat{\z}=(z_2,\ldots,z_d)\,$.
\noindent
By symmetry, the integral on the left hand side is a vector in the
same direction as $\z_0$.
We compute its magnitude using the angle
$\theta = \cos^{-1}(-z/\rho)$,
where $z$ is the component of $\z_\rho$
in the direction of $\z_0$.
%Using $S_{n-1}$ to denote the surface area of
%an $n$-dimensional sphere with unit radius,
The locus of points $\z_\rho\in\Srho$ with for which this angle
is between $\theta$ and $\theta\!+\!d\theta$
is an interval of length $\rho\,d\theta$ crossed with
%with a fixed value of $z$ is
a sphere of dimension $(d\!-\!2)$ with radius
$\rho\sin\theta$ and surface area
%$S_{d-2}(\rho\sin\theta)^{d-2}
%=S_{d-2}\rho^{d-2}(1-cos^2\theta)^{\frac{d-3}{2}}\sin\theta$.
$(2\pi^{\frac{d-1}{2}}\!/\Gamma(\frac{d-1}{2}))
(\rho\sin\theta)^{d-2}$.
For each point $\z_\rho$, the component of
$(\z_0-\z_\rho)$ in the direction of $\z_0$ is
$\z_0(1+\cos\theta)$.
Furthermore,
% $dz = \rho\sin(\theta)\,d\theta$ and
\[
||\z_0-\z_\rho||^2=4\rho^2\sin^2(\frac{\pi-\theta}{2})
=2\rho^2(1+\cos\theta).
\]
%Because $\Sphere$ is radially symmetric,
%the vector on the right hand side of Eqn(1)
%will share the same direction as $\,\z_i$,
%so we only need to consider
%the magnitude of the two vectors and,
%Without loss of generality, we can assume
%that only the first component of $\z_\rho$ is nonzero, i.e.
%$\z_\rho = (\rho,0,\ldots,0)$. 
Rewriting $(\sin\theta)^{d-2}$ as
$(1-\cos^2\theta)^{\frac{d-3}{2}}(\sin\theta)$ and
dividing by the surface area
$\rho^{d-1}2\pi^{\frac{d}{2}}/\Gamma(\frac{d}{2})$
for normalization, we have
\begin{align}
%\frac{1}{2d\!+\!1}
\int_{\Sphere_\rho}
\frac{2\mu(\z_0-\z_\rho)}{1+\frac{||\z_0-\z_\rho||^2}{\N}}\,d\z_\rho
&=\,\z_0\frac{\Gamma(\frac{d}{2})}{\Gamma(\!\frac{d-1}{2}\!)}
\!\int_{\theta=0}^{\;\pi}\!\!\frac{2\mu(1\!+\cos\theta)}{\sqrt{\pi}}
\frac{(1-\cos^2\theta)^{\frac{d-3}{2}}}{1+\frac{2\rho^2(1+cos(\theta))}{\N}}
\sin\theta\,d\theta \nonumber \\
&=\,\z_0\frac{\!2\mu\!}{\!\!\sqrt{\pi}}\frac{\Gamma(\frac{d}{2})}{\Gamma(\!\frac{d-1}{2}\!)}
\!\int_{\theta=0}^{\;\pi}\!\!
\frac{(1\!+\cos\theta)^{\frac{d-1}{2}}(1\!-\cos\theta)^{\frac{d-3}{2}}\!\!}
{1+\frac{2\rho^2(1+cos(\theta))}{\N}}
\sin\theta\,d\theta. \nonumber
\end{align}
We change variables to
$u = 1 + \frac{1+\cos\theta}{a}$, where
$a=\N/(2\rho^2)$, $\rho=\sqrt{\N/2a}\,$.$\;\;$Then
%\[
%u = 1 + \frac{1+\cos(\theta)}{a}, \quad\hbox{where}\quad
%a=\N/(2\rho^2),\qquad
%\rho=\sqrt{\N/2a}
%\]
\[
\cos\theta=
a(u-1) - 1,\quad
du=-\frac{1}{a}\sin\theta\,d\theta.
%au - b,\quad\hbox{where}\quad
%a=\frac{2d+1}{2\rho^2},\;\;b=\frac{2\rho^2+2d+1}{2\rho^2}
\]
So
\[
\int_{\Sphere_\rho}
\frac{2\mu(\z_0-\z_\rho)}{1+\frac{||\z_0-\z_\rho||^2}{\N}}\,d\z_\rho
=\,\z_0
\frac{2a\mu}{\sqrt{\pi}}
\frac{\Gamma(\frac{d}{2})}{\Gamma(\!\frac{d-1}{2}\!)}
\!\int_{u=1}^{1+\frac{2}{a}}
\!\frac{(a(u\!-\!1))^{\frac{d-1}{2}}(2-a(u\!-\!1))^{\frac{d-3}{2}}\!\!}{u}\,du\,.
\]

\medskip\noindent
Hence, the spherical distribution $\Sphere_\rho$ is stationary,
provided~the~expression~on~the~right~hand~side~is~equal~to~$\z_0$.
%\[
%\int_{\z_\rho}
%\frac{2(\z_0-\z_\rho)}{1+\frac{||\z_0-\z_\rho||^2}{2d+1}}
%=\z_0\frac{(2a)^{\frac{d+1}{2}}}{2}
%\frac{S_{d-2}}{S_{d-1}}
%\!\int_{u=1}^{1+\frac{2}{a}}
%\frac{(u-1)^{\frac{d-1}{2}}(1-\frac{a}{2}(u-1))^{\frac{d-3}{2}}}{u}du
%\]

\subsection{Finding an Approximate Value for $\N$}

If $d$ and $\mu$ are given, we would like to choose $\N$
in such a way that the radius $\rho$
of the stable spherical distribution
for $l_{\mu,\N}$
is very close to $\sqrt{d}\,$.
Let us define
\[
f_{d,a}(u) = 
\frac{2a}{\sqrt{\pi}}
\frac{\Gamma(\frac{d}{2})}{\Gamma(\frac{d-1}{2})}
\frac{(a(u-1))^{\frac{d-1}{2}}(2-a(u-1))^{\frac{d-3}{2}}}{u}\,,
\]
and let $a_0$ be the value of $a$ for which
$\int_{u=1}^{1+\frac{2}{a}}f_{d,a}(u)\,du = \frac{1}{\mu}$.
It follows from Theorem~1 that
$\rho=\sqrt{\N/2a_0}=\sqrt{d}$ when
$\N=2da_0$.
We can use the following Lemma to find a good approximation
$a_1$ for $a_0$ and hence a good choice
$\N=2da_1$ for $\N$.

\begin{lemma}
\mbox{}
\begin{itemize}
\item[(a)]
For $d \ge 3 $ and $0 < a < 2$,
\[
\int_{u=1}^{1+\frac{2}{a}}u\,f_{d,a}(u)\,du = 2\,.
\]
\item[(b)]
For $d\ge 4$,
the value $u_{(d,a)}$ of $u$ for which $f_{d,a}(u)$ is maximal
satisfies the equation
\[
a = \bigl(1+\frac{1}{u_{(d,a)}(d-3)+1}\bigr)/\bigl(u_{(d,a)}-1\bigr).
\]
\end{itemize}
\end{lemma}

\noindent
{\bf Proof:} (See below)

%We can use this Lemma to find a value of $a$
%which approximates the value for which
%$\int_{u=1}^{1+\frac{2}{a}}f_{d,a}(u)\,du=\frac{1}{\mu}$.
\bigskip\noindent
If we assume that $f_{d,a}$ is approximately
Gaussian, then the mean value of $\,u\,$ when
averaged with weighting $f(u)$ should be
approximately equal to the value $u_{(d,a)}$
at which $f(u)$ is maximal, i.e.
\[
\bigl(
\hbox{$\int_{u=1}^{1+\frac{2}{a}}$}
u\,f(u)\,du\bigr)/
\bigl(\hbox{$\int_{u=1}^{1+\frac{2}{a}}$}f(u)\,du\bigr)
\simeq u_{(d,a)}.
\]
If $\,a\,$ were chosen such that $u_{(d,a)}=2\mu$,
we would have
\[
\hbox{$\int_{u=1}^{1+\frac{2}{a}}$}f(u)\,du
\simeq \bigl(
\hbox{$\int_{u=1}^{1+\frac{2}{a}}$}u\,f(u)\,du\bigr)/u_{(d,a)}
=\frac{2}{2\mu}=\frac{1}{\mu}.
\]
From Part (b) of the Lemma, this corresponds to
\[
a \simeq \bigl(1+\frac{1}{2\mu(d-3)+1}\bigr)/(2\mu-1).
\]
In practice, the distribution is slightly skewed,
and the true mean is a bit larger than $u_{(d,a)}$.
In order to correct for this difference,
we replace $(2\mu(d-3)+1)$ with $2\mu(d-1)$ in the above formula,
giving us
\[
\N=2da\,\simeq\, 2da_1 = 2d(1+\frac{1}{2\mu(d-1)})/(2\mu-1).
\]

\noindent
{\bf Proof of Lemma:}

\smallskip\noindent
(a) We prove equivalently that
$F_{d,a}=G_{d,a}$
where
\begin{align}
F_{d,a}&=\int_{u=1}^{1+\frac{2}{a}}
(a(u-1))^{\frac{d-1}{2}}(2-a(u-1))^{\frac{d-3}{2}}\,du, \nonumber \\
G_{d,a}&=\frac{\sqrt{\pi}}{a}
\frac{\Gamma(\frac{d-1}{2})}{\Gamma(\frac{d}{2})}\,. \nonumber
\end{align}
%When $d$ is even,
%\[
%G_{d,a}=\frac{\sqrt{\pi}}{2a}
%\frac{\sqrt{\pi}(d-3)!!}{2^{\frac{d-2}{2}}(\frac{d-2}{2})!}
%=\frac{\pi}{2a}\frac{(d-3)!!}{(d-2)!!}
%\]
%%When $d$ is odd,
%\[
%G_{d,a}=\frac{\sqrt{\pi}}{2a}
%\frac{2^{\frac{d-1}{2}}(\frac{d-3}{2})!}{\sqrt{\pi}(d-2)!!}
%=\frac{\;\;(d-3)!!}
%{a(d-2)!!}
%\]

Change variables to
\[
a(u-1) = \frac{2}{t^2+1},\quad
du = \frac{\!\!-4t}{a(t^2+1)^2}\,dt.
\]

Then
%\begin{align}
\[
F_{d,a} \,=\, \frac{1}{a}\int_0^\infty
\!\Bigl(\frac{2}{t^2+1}\Bigr)\!^{\frac{d-1}{2}}\!
\Bigl(\frac{2t^2}{t^2+1}\Bigr)\!^{\frac{d-3}{2}}\!
\frac{4t}{(t^2+1)^2}\,dt
\,=\, \frac{2^d}{a}\int_0^\infty
\!\!\frac{t^{d-2}}{(t^2+1)^d}\,dt.
\]
%\end{align}

If we set
\[
J(k,n) = \int_0^\infty\!\frac{t^k}{(t^2+1)^n}\,dt,
\]

then $J(0,1) = \frac{\pi}{2}$, $J(1,n)=\frac{1}{2(n-1)}$
for $n\ge 2$, and we can derive these recurrence relations for
$n,k\ge 2\,$:
\[
J(k,n)=\frac{k-1}{2(n\!-\!1)}\,J(k\!-\!2,n\!-\!1),\quad
J(0,n) = \frac{2n\!-\!3}{2n\!-\!2}\,J(0,n-1).
\]

It follows that when $d\ge 2$ is even,
\begin{align}
F_{d,a}=\frac{2^d}{a}J(d-2,d)
&=\frac{2^d}{a}\,
\frac{(d-3)!!(\frac{d}{2})!}
{2^{\frac{d-2}{2}}(d-1)!}\;J(0,\hbox{$\frac{d+2}{2}$}) \nonumber \\
&=\frac{2^d}{a}\,
\frac{(d-3)!!(\frac{d}{2})!}
{2^{\frac{d-2}{2}}(d-1)!}
\;\frac{(d-1)!!}{d\,!!}\frac{\pi}{2} \nonumber \\
&=\frac{\pi}{a}\,
\frac{(d-3)!!\;2^{\frac{d}{2}}(\frac{d}{2})!(d-1)!!}
{(d-2)!!\;\;\;d\;\;\;(d-1)!} \nonumber \\
&=\frac{\pi}{a}\frac{(d-3)!!}{(d-2)!!}
= G_{d,a}\,. \nonumber
\end{align}

When $d\ge 3$ is odd,
\begin{align}
F_{d,a}=\frac{2^d}{a}J(d-2,d)
&=\frac{2^d}{a}\,
\frac{(d-3)!!(\frac{d+1}{2})!}
{2^{\frac{d-3}{2}}(d-1)!}\;J(1,\hbox{$\frac{d+3}{2}$}) \nonumber \\
&=\frac{2}{a}\,
\frac{(d-3)!!\;2^{\frac{d+1}{2}}(\frac{d+1}{2})!}
{(d-1)!\;\;\;\;2(\frac{d+1}{2})} \nonumber \\
&=\frac{2}{a}\,
\frac{(d-3)!!\;(d+1)!!}
{(d-1)!\;\;(d+1)} \nonumber \\
&=\frac{2}{a}\,
\frac{(d-3)!!}{(d-2)!!}
=G_{d,a}\,. \nonumber
\end{align}

\noindent
(b) When $d\ge 4$,
the derivative of $f_{d,a}(u)$ is
$C_{(d,a)}D_{(d,a)}$ where
\begin{align}
C_{(d,a)}&=\frac{-2a^2}{\sqrt{\pi}}
\frac{\Gamma(\frac{d}{2})}{\Gamma(\frac{d-1}{2})}
(a(u-1))^{\frac{d-3}{2}}
(2-a(u-1))^{\frac{d-5}{2}}\!, \nonumber \\
D_{(d,a)}&=(a(u-1)-1)(u(d-3)+1)-1\,. \nonumber
\end{align}

This derivative will be zero when $D_{(d,a)}=0$,
i.e.\ when
\[
a = \bigl(1+\frac{1}{u(d-3)+1}\bigr)/\bigl(u-1\bigr).
\]

%\newpage

%{\bf Networks trained on MNIST}:

\begin{minipage}{0.5\textwidth}

\begin{figure}[H]
\begin{center}
\vspace*{0.3cm}
\hspace*{1cm}
\begin{tikzpicture}[scale=0.7]
  \draw[black, thick] (-4,0) -- (4,0);
  \draw[black, thick] (0,0) -- (-2.4,3.2);
  \draw[blue, thick] (0,0) circle (4);
  \filldraw [red] (-2.4, 3.2) circle (0.1);
  \filldraw [red] (-2.4,-3.2) circle (0.1);
  \draw[red, thick, dotted] (-2.4,-3.2) -- (-2.4, 3.2);

  \node[text width=1cm] at ( 4.9,0) {\Large $\z_0$};
  \node[text width=1cm] at (-2.4,3.5) {\Large $\z_\rho$};
  \node[text width=1cm] at (-0.4,1.9) {\Large $\rho$};
  \node[text width=1cm] at (-1.55,-0.35) {\Large $z$};
  \node[text width=1cm] at (-0.4,0.5) {\Large $\theta$};
  %\node[text width=1cm] at (-2.8,-3.8) {\Large ${\cal S}_{\!\!\sqrt{\rho^2-z^2}}^{\;d-2}$};
  \node[text width=1cm] at (-3.4,-3.7) {\Large ${\cal S}_{\rho\sin\theta}^{\,d-2}$};
  \node[text width=1cm] at (3.0,3.6) {\Large ${\cal S}_{\rho}^{\,d-1}$};

\end{tikzpicture}
\vspace*{-0.3cm}
\end{center}
\caption{The integral of
$\nabla_{\z_0}K(\z_0,\z_\rho)$ over a
\hbox{$(d\!-\!1)$-dimensional} sphere
of radius $\rho$
can be calculated using a series of
$(d\!-\!2)$-dimensional
spheres of radius $\rho\sin\theta$.}
\label{fig:spherical_integral}
\end{figure}
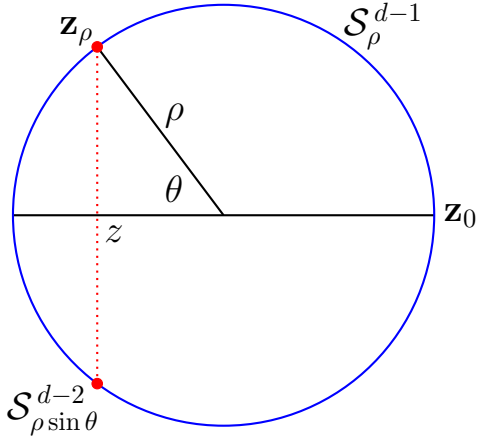

\smallskip

\subsection{Pytorch code for Eccentric Loss}

\smallskip

\hspace*{-0.6cm}
{\footnotesize\tt
\begin{tabular}{l}
\hline
 \\[-0.16cm]
$\!\!\!\!$def EccentricLoss($\,$z,$\,$scale=1$\,$): \\[0.1cm]
   x0$\;\;$= torch.squeeze(z) \\
   x1$\;\;$= x0.transpose(0,1) \\[0.1cm]
   batch$\,$=$\,$x0.size()[0] \\
   dim$\;\;\;$=$\,$x0.size()[1] \\
   norm$\,$=$\,$2*dim*(1 + 1/(2*scale*(dim-1)\hs)\hs)/(2*scale-1)$\!\!\!\!\!\!\!\!\!$ \\[0.1cm]
   xx$\;\;\,$=$\;$torch.bmm(x0.view(batch,1,dim), \\
   \phantom{xx$\;\;\,$=$\;$torch.bmm(}x0.view(batch,dim,1)\hs).squeeze(2)$\!\!\!\!\!\!\!\!\!$ \\
   xx0$\,$=$\;$xx.expand(batch,batch) \\
   xx1$\,$=$\;$xx0.transpose(0,1) \\[0.1cm]
   xy$\;\;$=$\;$xx0 + xx1 - 2*torch.matmul(x0,x1) \\[0.1cm]
   result$\,$=$\,$torch.sum(xx) - scale*norm* \textbackslash \\
\phantom{result$\,$=$\,$}torch.sum(torch.log(1+xy/norm)\hs)/batch \\[0.1cm]
   return result/(batch-1)  \\[0.2cm]
\hline
\end{tabular}
}

\vspace*{2cm}

\end{minipage}
\begin{minipage}{0.5\textwidth}

\subsection{Network Architectures}

\noindent
{\bf Encoder architecture for MNIST:}
{\small
\begin{align}
\!\!\!\!\!\!x\in{\cal R}^{32\times 32\times 1}
 & \rightarrow {\rm Conv}_{16}^{5(1)}\,\rightarrow{\rm BN}\rightarrow{\rm LeakyReLU}_{(0.1)}  \nonumber \\
 & \rightarrow {\rm Conv}_{24}^{4(2)}\,\rightarrow{\rm BN}\rightarrow{\rm LeakyReLU}_{(0.1)}  \nonumber \\
 & \rightarrow {\rm Conv}_{32}^{4(1)}\,\rightarrow{\rm BN}\rightarrow{\rm LeakyReLU}_{(0.1)}  \nonumber \\
 & \rightarrow {\rm Conv}_{48}^{4(2)}\,\rightarrow{\rm BN}\rightarrow{\rm LeakyReLU}_{(0.1)} \nonumber \\
 & \rightarrow{\rm FC}_{64} \rightarrow{\rm FC}_{d} \nonumber
\end{align}
}

\vspace*{-0.3cm}
\noindent
{\bf Decoder architecture for MNIST:}

{\small
\begin{align}
x\in{\cal R}^{d}
% & \rightarrow {\rm FC}_{8\times 8\times 1024}  \nonumber \\
 & \rightarrow {\rm FSConv}_{48}^{3(1)}\,\rightarrow{\rm BN}\rightarrow{\rm LeakyReLU}_{(0.1)}  \nonumber \\
 & \rightarrow {\rm FSConv}_{32}^{4(2)}\,\rightarrow{\rm BN}\rightarrow{\rm LeakyReLU}_{(0.1)}  \nonumber \\
 & \rightarrow {\rm FSConv}_{24}^{4(1)}\,\rightarrow{\rm BN}\rightarrow{\rm LeakyReLU}_{(0.1)}  \nonumber \\
 & \rightarrow {\rm FSConv}_{16}^{4(2)}\,\rightarrow{\rm BN}\rightarrow{\rm LeakyReLU}_{(0.1)}  \nonumber \\
 & \rightarrow {\rm FSConv}_{16}^{5(1)}\,\rightarrow{\rm BN}\rightarrow{\rm LeakyReLU}_{(0.1)}  \nonumber \\
 & \rightarrow {\rm FSConv}_{16}^{1(1)}\,\rightarrow{\rm BN}\rightarrow{\rm LeakyReLU}_{(0.1)}  \nonumber \\
 & \rightarrow {\rm FSConv}_{1}^{1}\;\;\;\rightarrow\;{\rm Sigmoid}  \nonumber
\end{align}
}

\vspace*{-0.2cm}
\noindent
{\bf Encoder architecture for CelebA:}

{\small
\begin{align}
\!\!\!\!\!\!\!\!x\in{\cal R}^{64\times 64\times 3}
 & \rightarrow {\rm Conv}_{128}^{4(2)}\,\rightarrow{\rm BN}\rightarrow{\rm ReLU}  \nonumber \\
 & \rightarrow {\rm Conv}_{256}^{4(2)}\,\rightarrow{\rm BN}\rightarrow{\rm ReLU}  \nonumber \\
 & \rightarrow {\rm Conv}_{512}^{4(2)}\,\rightarrow{\rm BN}\rightarrow{\rm ReLU}  \nonumber \\
 & \rightarrow {\rm Conv}_{1024}^{4(2)}\!\rightarrow{\rm BN}\rightarrow{\rm ReLU} \rightarrow{\rm FC}_{64}\!\!\!\!\!\!\!\!  \nonumber
\end{align}
}

\noindent
{\bf Decoder architecture for CelebA:}
{\small
\begin{align}
x\in{\cal R}^{64}
 & \rightarrow {\rm FC}_{8\times 8\times 1024}  \nonumber \\
 & \rightarrow {\rm FSConv}_{512}^{4(2)}\,\rightarrow{\rm BN}\rightarrow{\rm ReLU}  \nonumber \\
 & \rightarrow {\rm FSConv}_{256}^{4(2)}\,\rightarrow{\rm BN}\rightarrow{\rm ReLU}  \nonumber \\
 & \rightarrow {\rm FSConv}_{128}^{4(2)}\,\rightarrow{\rm BN}\rightarrow{\rm ReLU} \nonumber \\
 & \rightarrow{\rm FSConv}_{3}^{1}\;\;\;\rightarrow\;{\rm Sigmoid}  \nonumber
\end{align}
}

\end{minipage}

%{\bf Metaparameters for all networks:}

%\smallskip

%\begin{tabular}{ll}
%optimizer: & Adam \\
%batch size: & $100$ \\
%learning rate: & $0.0001$ \\
%weight decay: & $0.000001$ \\
%\end{tabular}

%\vfill

\end{document}